\documentclass[manuscript,screen]{acmart}
\usepackage{color,soul}
\usepackage{MnSymbol}
\usepackage{utfsym}
\usepackage{pifont}
\usepackage{color}
\usepackage{tabularray}
\usepackage{array}
\usepackage{colortbl}
\usepackage{multirow}
\usepackage{ragged2e}
\usepackage{booktabs}
\usepackage{hhline}
\usepackage{bbm}
\usepackage{rotating}
\usepackage{makecell}
\usepackage{wasysym}
\usepackage{booktabs,caption}
\usepackage[flushleft]{threeparttable}
\usepackage{xcolor}
\usepackage{enumitem}
\setlist[itemize]{noitemsep, topsep=0pt}
\usepackage[compact]{titlesec}
\usepackage{wrapfig}
\usepackage{tikz}
\newcommand*\circled[1]{\tikz[baseline=(char.base)]{
	\node[shape=circle,draw,fill=lightgray,text=black,inner sep=1pt] (char) {#1};}}
\usepackage{etoolbox}
\AtBeginEnvironment{tablenotes}{\footnotesize}

\titlespacing{\section}{0pt}{3ex}{1ex}

\AtBeginDocument{%
\providecommand\BibTeX{{%
		\normalfont B\kern-0.5em{\scshape i\kern-0.25em b}\kern-0.8em\TeX}}}

\setcopyright{acmlicensed}
\copyrightyear{2018}
\acmYear{2018}
\acmDOI{XXXXXXX.XXXXXXX}

\acmConference[Conference acronym 'XX]{Make sure to enter the correct
conference title from your rights confirmation emai}{June 03--05,
2018}{Woodstock, NY}
\acmISBN{978-1-4503-XXXX-X/18/06}

\begin{document}

\title{Passive Deepfake Detection: A Comprehensive Survey across Multi-modalities}

\author{Hong-Hanh Nguyen-Le}
\email{hong-hanh.nguyen-le@ucdconnect.ie}

\affiliation{%
	\institution{University College Dublin}
	\streetaddress{Belfield, Dublin 4}
	\city{Dublin}
	\country{Ireland}
	\postcode{D04V1W8}
}

\author{Van-Tuan Tran}
\affiliation{%
	\institution{Trinity College Dublin}
	\streetaddress{College Green, Dublin 2}
	\city{Dublin}
	\country{Ireland}
	\postcode{D02PN40}
}
\email{tranva@tcd.ie}

\author{Dinh-Thuc Nguyen}
\affiliation{%
	\institution{University of Science}
	\streetaddress{227 Nguyen Van Cu Street, Ward 4, District 5}
	\city{Ho Chi Minh City}
	\country{Vietnam}
}
\email{ndthuc@fit.hcmus.edu.vn}

\author{Nhien-An Le-Khac}
\affiliation{%
	\institution{University College Dublin}
	\streetaddress{Belfield, Dublin 4}
	\city{Dublin}
	\country{Ireland}
	\postcode{D04V1W8}
}
\email{an.lekhac@ucd.ie}

\renewcommand{\shortauthors}{H. Nguyen-Le, et al.}

\begin{abstract}
	In recent years, deepfakes (DFs) have been utilized for malicious purposes, such as individual impersonation, misinformation spreading, and artists' style imitation, raising questions about ethical and security concerns. In this survey, we provide a comprehensive review and comparison of passive DF detection across multiple modalities, including image, video, audio, and multi-modal, to explore the inter-modality relationships between them. Beyond detection accuracy, we extendour analysis to encompass crucial performance dimensions essential for real-world deployment: generalization capabilities across novel generation techniques, robustness against adversarial manipulations and postprocessing techniques, attribution precision in identifying generation sources, and resilience under real-world operational conditions. Additionally, we analyze advantages and limitations of existing datasets, benchmarks, and evaluation metrics for passive DF detection. Finally, we propose future research directions that address these unexplored and emerging issues in the field of passive DF detection. This survey offers researchers and practitioners a comprehensive resource for understanding the current landscape, methodological approaches, and promising future directions in this rapidly evolving field.
\end{abstract}



\received{20 February 2007}
\received[revised]{12 March 2009}
\received[accepted]{5 June 2009}

\maketitle

\section{Introduction}
The term \textbf{deepfake} \textit{(DF)} describes synthetic media (e.g., image, video, audio) produced by \textbf{generative models} (GMs) for malicious purposes, causing serious threats across financial, political, and social domains. In February 2024, a sophisticated DF attacks on Arup Group caused a $\$25.6$ million loss when the attacker used AI-generated avatars of CEO in a video conference \cite{Chen2024-os}. Similarly, in Singapore, a coordinated campaign impersonating political leaders defrauded $12,000$ investors of $\$180$ million through synthetic endorsements of cryptocurrency schemes \cite{Surashit-2024}. Beyond financial fraud, DF have been used for electoral interference where $23,000$ voters received AI-generated robocalls that mimic President Biden's voice patterns to suppress voter turnout \cite{Guardian-2024}.

In response to these emerging threats, two main categories of DF detection approaches have been developed: proactive and passive. Proactive approaches counteract DF manipulation before it occurs by adding perturbations to human data or embedding watermarks during the synthetic data generation process \cite{Nguyen_Le_2024}. In contrast, passive approaches are popular and straightforward solution against DFs as they are cast as binary classification problem. Passive approaches detect DFs after they are generated, focusing on analyzing intrinsic properties of them to distinguish between authentic and synthetic content. In this work, we focus on passive approaches rather than proactive ones.

\begin{table}[ht]
	\centering
	\fontsize{5pt}{5pt}\selectfont
	\caption{Comparative analysis of existing surveys on passive DF detection methods.}
	\label{tab:compare-survey}
	\begin{tblr}{
			width = \linewidth,
			colspec = {Q[250]Q[25]Q[100]Q[220]Q[220]},
			row{1} = {c},
			cell{2}{2} = {c},
			cell{2}{3} = {c},
			cell{3}{2} = {c},
			cell{3}{3} = {c},
			cell{4}{2} = {c},
			cell{4}{3} = {c},
			cell{5}{2} = {c},
			cell{5}{3} = {c},
			cell{6}{2} = {c},
			cell{6}{3} = {c},
			cell{7}{2} = {c},
			cell{7}{3} = {c},
			cell{8}{2} = {c},
			cell{8}{3} = {c},
			cell{9}{2} = {c},
			cell{9}{3} = {c},
			cell{10}{2} = {c},
			cell{10}{3} = {c},
			cell{11}{2} = {c},
			cell{11}{3} = {c},
			cell{12}{2} = {c},
			cell{12}{3} = {c},
			cell{12}{5} = {c},
			vline{2-5} = {-}{},
			hline{1,13} = {-}{0.08em},
			hline{2,12} = {-}{},
			hline{2} = {2}{-}{},
		}
		\textbf{Survey Papers}                                                                                                             & \textbf{Year} & \textbf{Modality}                          & \textbf{Strengths}                                                                                                                                                                                                                                                     & \textbf{Limitations}                                                                       \\
		The creation and detection of deepfakes: A survey \cite{mirsky2021creation}                                                        & 2021          & Image                                    & In-depth explanations of GAN-based generation techniques                                                                                                                                                                                                               & Lack discussion on DF detection challenges                                                 \\
		Deepfake detection for human face images and videos: A survey \cite{malik2022deepfake}                                             & 2022          & Image \& Video                             & Categories of DF facial generation techniques                                                                                                                                                                                                                          & Lacks analysis of cross-dataset generalization challenges                                  \\
		Gan-generated faces detection: A survey and new perspectives~\cite{wang2202gan}                                                    & 2022          & Image                                    & Specialized focus on GAN-specific artifacts                                                                                                                                                                                                                            & Excludes non-GAN synthetic media                                                           \\
		Audio deepfake detection: A survey \cite{yi2023audio}                                                                              & 2023          & Audio                                    & Comprehensive overview of backbones and feature extraction techniques for audio DF detection                                                                                                                                                                           & Lack discussion on DF audio detection challenges                                           \\
		Deepfakes generation and detection: state-of-the-art, open challenges, countermeasures, and way forward \cite{masood2023deepfakes} & 2024          & Single-modal                             & Extensive survey about generation and detection techniques with 300+ references                                                                                                                                                                                        & Only focus on discussion on detection accuracy                                             \\
		A Survey on the Detection and Impacts of Deepfakes in Visual, Audio, and Textual Formats \cite{mubarak2023survey}                  & 2024          & Single-modal                             & Discussion on impacts of DFs on polity, society, and economy aspects                                                                                                                                                                                                   & Simple categories: methods using handcreafted features and methods using DL                \\
		Deepfake Generation and Detection: A Benchmark and Survey \cite{pei2024deepfake}                                                   & 2024          & Image \& Video                             & Benchmarks for~evaluating GM models                                                                                                                                                                                                                                    & Limited discussion of adversarial robustness in detection models                           \\
		Deepfake video detection: challenges and opportunities \cite{kaur2024deepfake}                                                     & 2024          & Video                                    & Comprehensive survey of real-time applications of DF video detection                                                                                                                                                                                                   & No discussion on generalization and robustness aspects of real-world applications          \\
		Evolving from Single-modal to Multimodal Facial Deepfake Detection: A Survey \cite{liu2024evolving}                                & 2024          & Single- and Multimodal in image and video modalities~ & Review both passive and proactive DF detection                                                                                                                                                                                                                         & No discussion on robustness, generalization, real-world resilience and attribution aspects \\
		Deepfake detection: A comprehensive study from the reliability perspective \cite{wang2022deepfake}                                 & 2024          & Image                                    & Empirical analysis of detection reliability                                                                                                                                                                                                                            & No detailed discussion about passive approaches                                            \\
		Our Survey                                                                                                                         & 2025          & Single- and multimodal across modalities    & {- Comprehensive categories of DF detection across multi-modalities and analysis the relationship between them;\\- Discussion beyond detection accuracy: generalization, robustness, attribution, and real-world resilience\\- Analysis of limitations of DF datasets} & -
	\end{tblr}
	\vspace*{-2.7\baselineskip}
\end{table}

\subsection{Motivation}
While there are several surveys of passive DF detection for specific modality \cite{kaur2024deepfake, mirsky2021creation, wang2202gan, yi2023audio, malik2022deepfake}, several critical aspects motivate a comprehensive survey of this field.

First, there is a pressing need to understand the relationships between DF detection approaches across different modalities. While image-based detection methods have shown promising results, their direct application to video or audio modalities may be suboptimal due to the unique temporal characteristics \cite{liu2023ti2net,gu2021spatiotemporal,yin2023dynamic} and modality-specific artifacts present in the audio modality \cite{sun2023ai,yan2022initial}. For instance, methods that excel at detecting facial manipulations in static images often struggle with temporal inconsistencies in videos. Similarly, single-modal approaches prove insufficient for detecting sophisticated talking-face video generations \cite{xu2024vasa,tan2024say}, where both visual and audio streams are manipulated \cite{cozzolino2023audio,feng2023self}. Therefore, it is crucial to understand these \textbf{inter-modality relationships} for developing effective detection approaches that can leverage complementary information across modalities.

Second, when deploying DF detection systems in real-world settings, focusing \textbf{solely} on detection accuracy becomes \textbf{inadequate}. Research has increasingly recognized this limitation, leading to the development of specialized techniques addressing several critical enhancement dimensions:
\begin{itemize}
	\item \textbf{Generalization}: Several works demonstrate that detection methods perform well when training and test samples come from the same dataset but struggle with distribution shifts during inference \cite{corvi2023intriguing, ojha2023towards, corvi2023detection, chen2022ost}. This problem raises needs to develop dedicated methods to maintain detection reliability across various datasets, manipulation techniques, and generator architectures.
	\item \textbf{Robustness}: Detectors show vulnerability to adversarial attacks and postprocessing attacks \cite{corvi2023intriguing, hou2023evading}, which degrade the accuracy performance of detectors. It is crucial to develop methods to enhance detectors' robustness to these attacks.
	\item \textbf{Attribution}: Beyond binary classification, identifying the specific generator source of DFs becomes crucial for forensic analysis and accountability. This capability allows law enforcement and cybersecurity professionals to establish connections between different instances of fake content and track the activities of bad actors \cite{xie2024generalized, asnani2023reverse, yang2022deepfake}.
	\item \textbf{Real-world resilience}: In real-world settings, detectors encounter naturally occurring conditions, including DF quality degradation caused by compression algorithms of social media platforms or messaging applications, variations in media sizes and temporal synchronization issues \cite{feng2023self, xu2024learning, le2023quality}. Maintaining reliable performance under these conditions is also important for DF detectors.
\end{itemize}

These motivations underscore the need for a comprehensive survey that not only examines detection approaches across different modalities but also reviews the substantial body of work specifically developed to enhance generalization, robustness, attribution, and real-world resilience of DF detection systems.

\subsection{Related survey work}

Previous surveys have typically focused on specific modalities in passive DF detection, including image \cite{mirsky2021creation,malik2022deepfake,wang2202gan,pei2024deepfake,wang2022deepfake}, video \cite{malik2022deepfake,pei2024deepfake,kaur2024deepfake}, and audio \cite{yi2023audio}. Table \ref{tab:compare-survey} compares our work with existing survey papers, highlighting the strengths and drawbacks of each work.

\citeauthor{mirsky2021creation} \cite{mirsky2021creation} provides comprehensive insights into GAN architectures employed in DF generation and \citeauthor{pei2024deepfake} \cite{pei2024deepfake} focused more on benchmarks for evaluating GMs. \citeauthor{wang2202gan} \cite{wang2202gan} reviews detection techniques specifically for identifying GAN-generated artifacts, encompassing DL-based, physical-based, and physiological-based methods, while \citeauthor{malik2022deepfake} \cite{malik2022deepfake} provides broader categories of DF detection approaches in both image and video modalities. \cite{yi2023audio} is the first work that reviews existing approaches for detecting fake audios. Instead of reviewing state-of-the-art (SoTA) detection approaches, \citeauthor{kaur2024deepfake} examines real-world applications of DF video detectors, particularly focusing on computational complexity and scalability considerations. Additionally, \citeauthor{wang2022deepfake} contributes valuable insights into the reliability aspects of image-based DF detection methods. Recent surveys have reviewed single-modal DF detection approaches \cite{masood2023deepfakes, mubarak2023survey} or both single- and multi-model approaches \cite{liu2024evolving}.

In contrast to previous surveys, our work presents several distinctive characteristics:
\begin{itemize}
	\item Rather than focusing on a specific modality, our work presents a comprehensive cross-modality survey that examines both single- and multimodal approaches across modalities. We also analyze the relationships of detection approaches between modalitys.
	\item Previous works only focus on methods that improve detection accuracy, leaving critical aspects of real-world deployment, such as generalization and robustness, relatively unexplored. In this work, we review methods that aim to develop DF detectors to satisfy other practical aspects, including generalization, robustness, attribution, and real-world resilience. We also analyze trade-offs about efficiency and computation if DF detectors can satisfy all these aspects.
\end{itemize}

\subsection{Contributions}
In summary, the main contributions of our work are as follows:
\begin{itemize}
	\item A comprehensive cross-modality review of passive DF detection approaches, examining relationships between techniques across image, video, audio, and multimodal modalities.
	\item An extended evaluation framework that goes beyond detection accuracy to address critical deployment considerations including generalization capacity, adversarial robustness, attribution capabilities, and resilience in real-world scenarios.
	\item An analysis of limitations in current benchmark datasets, highlighting gaps that affect the development and evaluation of detection systems.
	\item Insights into  emerging challenges and potential implementations for future research.
\end{itemize}

\subsection{Survey methodology}
This survey follows a systematic approach to select relevant literature on passive DF detection.

\textbf{Publication venues}. We have conducted an exhaustive search for relevant publications published in top-tier artificial intelligence (AI) and security venues, including AI/security conferences and leading journals. We use databases such as IEEE Xplore, ACM Digital Library, and Google Scholar.


\begin{wraptable}{r}{6.0cm}
	\centering
	\fontsize{5pt}{5pt}\selectfont
	\caption{Numerical comparison between reported results and DeepfakeBench's results}
	\label{tab:numerical-compare}
	\begin{tblr}{
			width = \linewidth,
			colspec = {Q[170]Q[75]Q[95]Q[95]Q[75]Q[95]Q[95]},
			cells = {c},
			cell{1}{1} = {r=2}{},
			cell{1}{2} = {c=3}{0.25\linewidth},
			cell{1}{5} = {c=3}{0.35\linewidth},
			vline{2,5} = {1-8}{},
			hline{1,9} = {-}{0.08em},
			hline{2} = {2-7}{},
			hline{3} = {-}{},
			hline{3} = {2}{-}{},
		}
		\textbf{Detector }            & \textbf{Reported results } &               &                  & \textbf{DeepfakeBench's Results } &               &                  \\
		& \textbf{FF++}              & \textbf{DFDC} & \textbf{CelebDF} & \textbf{FF++}                     & \textbf{DFDC} & \textbf{CelebDF} \\
		F3Net \cite{qian2020thinking} & 98.62                      & -             & -                & 97.93                             & -             & -                \\
		SRM \cite{liu2021spatial}     & 96.9                       & 79.7          & 79.4             & 93.59                             & 69.95         & 75.52            \\
		X-ray \cite{li2020face}       & 98.52                      & 80.92         & 80.58            & 93.91                             & 63.26         & 67.86            \\
		Recce \cite{cao2022end}       & 95.02                      & 69.06         & 99.94            & 81.90                             & 71.33         & 73.19            \\
		Core \cite{ni2022core}        & 99.94                      & 72.41         & 75.71            & 94.31                             & 73.41         & 74.28            \\
		UCF \cite{yan2023ucf}         & 99.6                       & 80.5          & 82.4             & 95.27                             & 71.91         & 75.27
	\end{tblr}
	\vspace*{-0.7\baselineskip}
\end{wraptable}

\textbf{Comparison strategy}. We take a conceptual rather than a numerical approach. As indicated by DeepfakeBench \cite{yan2023deepfakebench}, previously reported results in DF detection papers may be biased or misleading due to the inconsistent experimental settings and the lack of standardization across papers in terms of evaluation strategies and metrics. Therefore, instead of comparing methods based on reported numerical results, we analyze and compare approaches based on their underlying concepts, methodologies, and key contributions. This conceptual comparison provides a more meaningful and reliable assessment of different detection approaches while avoiding the pitfalls of direct numerical comparisons across inconsistent evaluation frameworks. Table \ref{tab:numerical-compare} shows the inconsistent results.

\begin{figure}[ht]
	\centering
	\includegraphics[width=0.99\linewidth]{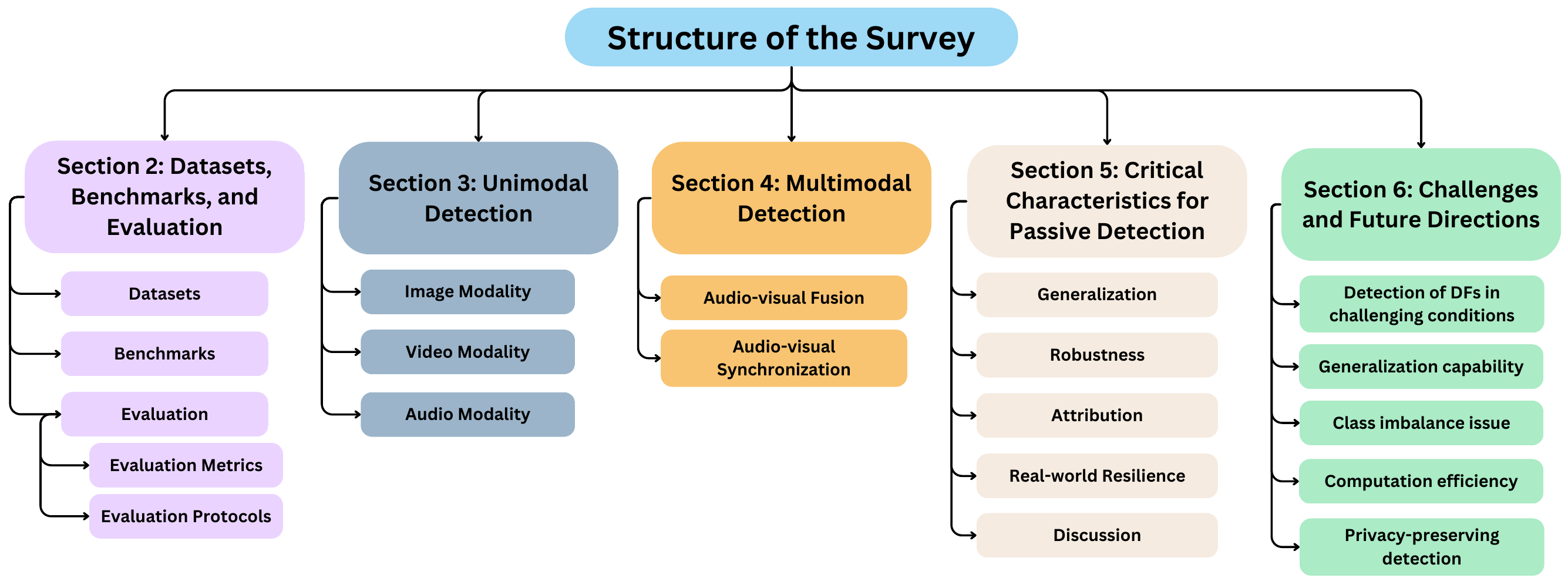}
	\caption{Overview of our survey structure.}
	\label{fig:structure}
	\vspace*{-1.7\baselineskip}
\end{figure}

The rest of this survey is structured as follows: Section \ref{sec:data-metric} presents common benchmarks, datasets, and evaluation metrics. We review unimodal approaches in Section \ref{sec:unimodal} and multimodal approaches in Section \ref{sec:multimodal}. In Section \ref{sec:beyond}, we explores aspects beyond detection accuracy, including generalization, robustness, attribution, and real-world resilience. Section \ref{sec:challenge-future} discusses current challenges and future research directions and Section \ref{sec:conclusion} concludes the survey with key findings and insights. Figure \ref{fig:structure} provides an overview of our survey organization.

\section{Datasets, Benchmarks and Evaluation}\label{sec:data-metric}

\subsection{Datasets} \label{subsec:dataset}
Visual modality datasets (both image and video) dominate the field in terms of volume and diversity, with over 20 publicly available datasets \cite{dang2020detection, he2021forgerynet, le2021openforensics, wang2023dire, yan2024df40, yang2019exposing, korshunov2018deepfakes, rossler2019faceforensics++, dolhansky2019deepfake, li2020celeb, jiang2020deeperforensics, zi2020wilddeepfake, he2021forgerynet, zhou2021face, kwon2021kodf, narayan2022deephy, narayan2023df} compared to only 11 for audio \cite{wang2020asvspoof, liu2023asvspoof, frank2021wavefake, yi2022add, yi2023add, li2024cross, muller2022does, yaroshchuk2023open, zhang2022partialspoof, muller2024mlaad, yan2024voicewukong} and 3 for multimodal content \cite{khalid2021fakeavceleb, cai2022you, cai2023av}. More than $90\%$ of research papers trained their models on FF++ \cite{rossler2019faceforensics++} in image and video modalities and on WaveFake \cite{frank2021wavefake} or ASVspoof2019 \cite{wang2020asvspoof} in the audio modality. To evaluate generalization capabilities, researchers frequently employ datasets containing real-world DFs like CelebDF \cite{li2020celeb}, DFDC \cite{dolhansky2019deepfake}, ForgeryNet \cite{he2021forgerynet}, DF-1.0 \cite{jiang2020deeperforensics}, Wild-DF \cite{zi2020wilddeepfake}, ADD \cite{yi2023add}, and FakeAVCeleb \cite{khalid2021fakeavceleb} as these better represent the challenges of detecting DFs in-the-wild, such as unknown synthesis models or low fake quality. Recent datasets have begun incorporating new generator architectures, particularly diffusion models (DMs). In the visual modality, DF40 \cite{yan2024df40} represents a significant advancement by comprising 40 distinct SoTA synthesis models, including both GANs and DMs, along with 4 different manipulation techniques. Similarly, in the audio modality, VoiceWukong \cite{yan2024df40} and MLAAD \cite{muller2024mlaad} stand out by covering 34 and 82 synthesis models, respectively, from both research and commercial sources.

However, there are critical limitations across all modalities: most datasets inadequately capture the dynamic and unpredictable nature of real-world scenarios and suffer from class imbalance. As noted by \citeauthor{layton2024sok} \cite{layton2024sok}, most datasets suffer from significant class imbalance, leading to biased training and poor real-world performance. VoiceWukong \cite{yan2024df40} is the only dataset that maintains a balanced distribution between real and fake samples.
Moreover, only a few datasets like OpenForensics \cite{le2021openforensics}, DF-Platter \cite{narayan2023df}, and DeeShy \cite{narayan2022deephy} attempt to involve challenging scenarios such as multiple fake faces in an image or video, low-quality DFs, or face occlusion. Table \ref{tab:passive-dataset} summarizes datasets on passive DF detection for each modality.

\begin{table}[ht]
	\centering
	\fontsize{5pt}{5pt}\selectfont
	\caption{A Summarization of Datasets for Each Modality in Passive DF Detection Approaches.}
	\label{tab:passive-dataset}
	\begin{tblr}{
			width = \linewidth,
			colspec = {Q[20]Q[160]Q[30]Q[40]Q[45]Q[25]Q[25]Q[55]Q[90]Q[50]Q[75]Q[65]Q[65]Q[40]Q[45]},
			cells = {c},
			cell{1}{1} = {r=2}{},
			cell{1}{2} = {r=2}{},
			cell{1}{3} = {r=2}{},
			cell{1}{4} = {c=2}{0.05\linewidth},
			cell{1}{6} = {c=2}{0.05\linewidth},
			cell{1}{8} = {r=2}{},
			cell{1}{9} = {r=2}{},
			cell{1}{10} = {r=2}{},
			cell{1}{11} = {r=2}{},
			cell{1}{12} = {r=2}{},
			cell{1}{13} = {r=2}{},
			cell{1}{14} = {r=2}{},
			cell{3}{1} = {r=6}{},
			cell{9}{1} = {r=14}{},
			cell{23}{1} = {r=11}{},
			cell{34}{1} = {r=3}{},
			vline{2-4,6,8-14} = {1-36}{},
			hline{1,37} = {-}{0.08em},
			hline{2} = {4-7}{},
			hline{3,9,23,34} = {-}{},
			hline{3} = {2}{-}{},
		}
		& \textbf{Name}                              & \textbf{Year} & \textbf{Samples } &               & \textbf{Generator} &              & \textbf{Synthesis model} & \textbf{Manipulation Technique} & \textbf{Multiple faces} & \textbf{Low quality} & \textbf{Face occlusion} & \textbf{Language} & \textbf{Partial fake} \\
		&                                            &               & \textbf{Real}     & \textbf{Fake} & \textbf{GANs}      & \textbf{DMs} &                          &                                 &                         &                      &                         &                   &                       \\
		\begin{sideways}Image\end{sideways}       & DFFD \cite{dang2020detection}              & 2019          & 58,703            & 240,336       & \usym{2714}        & \usym{2718}  & 7                        & FS, FR, FE                      & \usym{2718}             & \usym{2718}          & \usym{2718}             & -                 & \usym{2718}           \\
		& ForgeryNet \cite{he2021forgerynet}         & 2021          & 1,438,201         & 1,457,861     & \usym{2714}        & \usym{2718}  & 15                       & FS, FR, FE                      & \usym{2718}             & \usym{2718}          & \usym{2718}             & -                 & \usym{2718}           \\
		& OpenForensics \cite{le2021openforensics}   & 2021          & 160,467           & 173,660       & \usym{2714}        & \usym{2718}  & 1                        & FS                              & \usym{2714}             & \usym{2718}          & \usym{2714}             & -                 & \usym{2718}           \\
		& DiffusionForensics \cite{wang2023dire}     & 2023          & 134,000           & 137,200       & \usym{2718}        & \usym{2714}  & 9                        & ES                              & \usym{2718}             & \usym{2718}          & \usym{2718}             & -                 & \usym{2718}           \\
		& DiffusionFace \cite{chen2024diffusionface} & 2024          & 30,000            & 600,000       & \usym{2718}        & \usym{2714}  & 11                       & FS, ES, FE                      & \usym{2718}             & \usym{2718}          & \usym{2718}             & -                 & \usym{2718}           \\
		& DF40 \cite{yan2024df40}                    & 2024          & 1590              & 1M            & \usym{2714}        & \usym{2714}  & 40                       & ES, FE                          & \usym{2718}             & \usym{2718}          & \usym{2718}             & -                 & \usym{2718}           \\
		\begin{sideways}Video\end{sideways}       & UADFV \cite{yang2019exposing}              & 2018          & 49                & 49            & \usym{2714}        & \usym{2718}  & 1                        & FS                              & \usym{2718}             & \usym{2718}          & \usym{2718}             & -                 & \usym{2718}           \\
		& DF-TIMIT \cite{korshunov2018deepfakes}     & 2018          & 320               & 640           & \usym{2714}        & \usym{2718}  & 2                        & FS                              & \usym{2718}             & \usym{2714}          & \usym{2718}             & -                 & \usym{2718}           \\
		& DFFD \cite{dang2020detection}              & 2019          & 1,000             & 3,000         & \usym{2714}        & \usym{2718}  & 7                        & FS, FR, FE                      & \usym{2718}             & \usym{2718}          & \usym{2718}             & -                 & \usym{2718}           \\
		& FF++ \cite{rossler2019faceforensics++}     & 2019          & 1,000             & 4,000         & \usym{2714}        & \usym{2718}  & 4                        & FS, FR                          & \usym{2718}             & \usym{2718}          & \usym{2718}             & -                 & \usym{2718}           \\
		& DFDC \cite{dolhansky2019deepfake}          & 2019          & 1,131             & 4,113         & \usym{2714}        & \usym{2718}  & 8                        & -                               & \usym{2718}             & \usym{2718}          & \usym{2718}             & -                 & \usym{2718}           \\
		& Celeb-DF \cite{li2020celeb}                & 2020          & 590               & 5,639         & \usym{2714}        & \usym{2718}  & 1                        & FS, FR                          & \usym{2718}             & \usym{2718}          & \usym{2718}             & -                 & \usym{2718}           \\
		& DF-1.0 \cite{jiang2020deeperforensics}     & 2020          & 50,000            & 10,000        & \usym{2714}        & \usym{2718}  & 1                        & FS                              & \usym{2718}             & \usym{2718}          & \usym{2718}             & -                 & \usym{2718}           \\
		& Wild-DF \cite{zi2020wilddeepfake}          & 2021          & 3,805             & 3,509         & \usym{2714}        & \usym{2718}  & 1                        & -                               & \usym{2718}             & \usym{2718}          & \usym{2718}             & -                 & \usym{2718}           \\
		& ForgeryNet \cite{he2021forgerynet}         & 2021          & 99,630            & 121,617       & \usym{2714}        & \usym{2718}  & 15                       & FS, FR, FE                      & \usym{2718}             & \usym{2718}          & \usym{2718}             & -                 & \usym{2718}           \\
		& FFIW \cite{zhou2021face}                   & 2021          & 10,000            & 10,000        & \usym{2714}        & \usym{2718}  & 3                        & FS                              & \usym{2718}             & \usym{2718}          & \usym{2718}             & -                 & \usym{2718}           \\
		& KoDF \cite{kwon2021kodf}                   & 2021          & 62,166            & 175.776       & \usym{2714}        & \usym{2718}  & 6                        & FS, FR                          & \usym{2718}             & \usym{2718}          & \usym{2718}             & -                 & \usym{2718}           \\
		& DeeShy \cite{narayan2022deephy}            & 2022          & 100               & 5,040         & \usym{2714}        & \usym{2718}  & 3                        & FS, FR                          & \usym{2718}             & \usym{2718}          & \usym{2714}             & -                 & \usym{2718}           \\
		& DF-Platter \cite{narayan2023df}            & 2023          & 133,260           & 132,496       & \usym{2714}        & \usym{2718}  & 3                        & FS                              & \usym{2714}             & \usym{2714}          & \usym{2714}             & -                 & \usym{2718}           \\
		& DF40 \cite{yan2024df40}                    & 2024          & 1590              & 0.1M          & \usym{2714}        & \usym{2714}  & 40                       & FS, FR                          & \usym{2718}             & \usym{2718}          & \usym{2718}             & -                 & \usym{2718}           \\
		\begin{sideways}Audio\end{sideways}       & ASVspoof 2019 \cite{wang2020asvspoof}      & 2019          & 41,913            & 300,678       & -                  & -            & -                        & VC, TTS~                        & -                       & \usym{2718}          & -                       & en                & \usym{2718}           \\
		& ASVspoof 2021 \cite{liu2023asvspoof}       & 2021          & 22,617            & 589,212       & -                  & -            & -                        & VC, TTS                         & -                       & \usym{2718}          & -                       & en                & \usym{2718}           \\
		& WaveFake \cite{frank2021wavefake}          & 2021          & 18,100            & 117,985       & \usym{2714}        & \usym{2718}  & 6                        & VC                              & -                       & \usym{2718}          & -                       & en, jp            & \usym{2718}           \\
		& ADD 2022 \cite{yi2022add}                  & 2022          & 36,953            & 123,932       & -                  & -            & -                        & VC                              & -                       & \usym{2714}          & -                       & ch                & \usym{2714}           \\
		& ADD 2023 \cite{yi2023add}                  & 2023          & 172,819           & 113,042       & -                  & -            & -                        & VC                              & -                       & \usym{2714}          & -                       & ch                & \usym{2714}           \\
		& ITW \cite{muller2022does}                  & 2022          & 17,000            & 14,000        & -                  & -            & -                        & VC                              & -                       & \usym{2718}          & -                       & en                & \usym{2718}           \\
		& ODSS \cite{yaroshchuk2023open}             & 2023          & 11,032            & 18,993        & \usym{2714}        & \usym{2718}  & 2                        & TTS                             & -                       & \usym{2718}          & -                       & en, es, de        & \usym{2718}           \\
		& PS \cite{zhang2022partialspoof}            & 2023          & 12,483            & 121,461       & -                  & -            & -                        & VC, TTS                         & -                       & \usym{2718}          & -                       & en                & \usym{2714}           \\
		& CD-ADD \cite{li2024cross}                  & 2024          & 28,212            & 120,459       & \usym{2714}        & \usym{2714}  & 5                        & VC, TTS                         & -                       & \usym{2718}          & -                       & en                & \usym{2718}           \\
		& MLAAD \cite{muller2024mlaad}               & 2024          & 20,000            & 134,000       & \usym{2714}        & \usym{2714}  & 82                       & TTS                             & -                       & \usym{2718}          & -                       & many (38)         & \usym{2718}           \\
		& VoiceWukong \cite{yan2024voicewukong}      & 2024          & 413,400           & 413,400       & \usym{2714}        & \usym{2714}  & 34                       & VC, TTS                         & -                       & \usym{2714}          & -                       & en, ch            & \usym{2718}           \\
		\begin{sideways}multimodal\end{sideways} & FakeAVCeleb \cite{khalid2021fakeavceleb}   & 2022          & 500               & 19,500        & \usym{2714}        & \usym{2718}  & 4                        & FR, TTS                         & \usym{2718}             & \usym{2718}          & \usym{2718}             & en                & \usym{2718}           \\
		& LAV-DF \cite{cai2022you}                   & 2022          & 36,431            & 99,873        & \usym{2714}        & \usym{2718}  & 1                        & FR, TTS                         & \usym{2718}             & \usym{2718}          & \usym{2718}             & en                & \usym{2718}           \\
		& AV-Deepfake1M \cite{cai2023av}             & 2024          & 286,721           & 860,039       & \usym{2714}        & \usym{2718}  & 3                        & FR, TTS                         & \usym{2718}             & \usym{2718}          & \usym{2718}             & en                & \usym{2718}
	\end{tblr}
	\begin{tablenotes}
		\item i) \textbf{Generator}: Type of GMs that the dataset cover: GANs (Generative Adversarial Networks), DMs (Diffusion models)
		\item ii) \textbf{Synthesis Models}: The number of different synthesis models used to generate DFs in each dataset.
		\item iii) \textbf{Manipulation Techniques}: Techniques used for manipulation, including: face swapping (FS), face reenactment (FR), entirely face synthesis (ES), face editting (FE), partially fake audio (PF), voice conversion (VC), and text-to-speech (TTS).
	\end{tablenotes}
	\vspace*{-2.3\baselineskip}
\end{table}

\subsection{Benchmarks} \label{subsec:benchmark}
DeepfakeBench \cite{yan2023deepfakebench} is the first comprehensive and public benchmark for DF detection in image and video modalities, offering an integrated framework of 34 detectors and 10 datasets, multi-GPUs training, standardized data preprocessing, and evaluation protocols. Regarding the audio modality, VoiceWukong \cite{yan2024voicewukong} is the first comprehensive benchmark for DF detection which support 12 SoTA detectors. However, these benchmark do not cover datasets for challenging scenratios, such as multiple fake faces, face occlusions, partial fake, and low quality. Recently, \cite{li2023continual} introduces a continual benchmark that simulates a diverse stream of DFs sequentially over time, challenging detection models to learn new fake types without forgetting previously encountered ones. Despite its innovative approach to testing detector robustness against catastrophic forgetting, CDDB has received limited adoption in subsequent research.

\subsection{Evaluation} \label{ssec:evaluation}
\subsubsection{Evaluation Metrics.}
Metrics commonly used for DF detection evaluation include accuracy (ACC), area under the ROC curve (AUC), average precision (AP), F1-score, and equal error rate (EER). For detection methods that identify manipulated regions, intersection over union (IoU) is used to measure the overlap between predicted and ground-truth manipulation masks. Note that for the video modality, these metrics can be calculated either at the frame level or aggregated over the entire video sequence, depending on the specific task and evaluation criteria. However, due to the imbalanced nature of DF datasets \cite{layton2024sok}, where fake samples often outnumber real ones, solely using ACC for evaluation is not enough. A model could achieve high accuracy by simply classifying all samples as fake, yet fail to effectively distinguish fake content. Therefore, it is crucial to evaluate DF detection approaches using multiple metrics, particularly AUC and F1-score.

\subsubsection{Evaluation Protocols.}
Detection approaches are evaluated under different distinct protocols to assess their effectiveness, robustness, and generalization capabilities \cite{yan2023deepfakebench, yan2024df40}.

\circled{1} \textbf{Within-domain evaluation} examines the accuracy performance of the detector when training and testing on the same dataset. This protocol aims to assess the detection capability under controlled conditions where the manipulation techniques and data distributions are consistent. \circled{2} \textbf{Cross-domain evaluation} involves evaluating the detector's generalization capability across different/new datasets to handle the distribution shifts in the real-world settings. \circled{3} \textbf{Cross-manipulation evaluation} evaluates the detector's generalization capability across different manipulation methods (within the dataset). One common setting is that the detector trained on Face2Face in FF++, and tested on the remaining types of forged face in FF++. \circled{4} \textbf{Unknown-domain evaluation} presents the most challenging scenario, where test samples come from entirely unseen manipulation techniques, generators, and data distributions. For example, a detector trained on face swapping samples from FF++ might be tested on face reenactment examples from DF40. This protocol most closely simulates real-world deployment conditions, where detectors must identify DFs generated by novel, previously unseen techniques. \circled{5} \textbf{Adversarial attacks evaluation} measures the detector's resilience to adversarial examples (AEs). There are three common levels of attacks: (i) white-box attack - attackers have complete knowledge of the model architecture and parameters; (ii) black-box attack - attackers have access only to model outputs; and (iii) transferable attack - the most challenging attack in which attackers can create AEs without needing direct access to the ultimate target model. More details about different types of adversarial attacks can be explored in \cite{li2024survey}.

\begin{figure}[ht]
	\centering
	\includegraphics[width=0.9\linewidth]{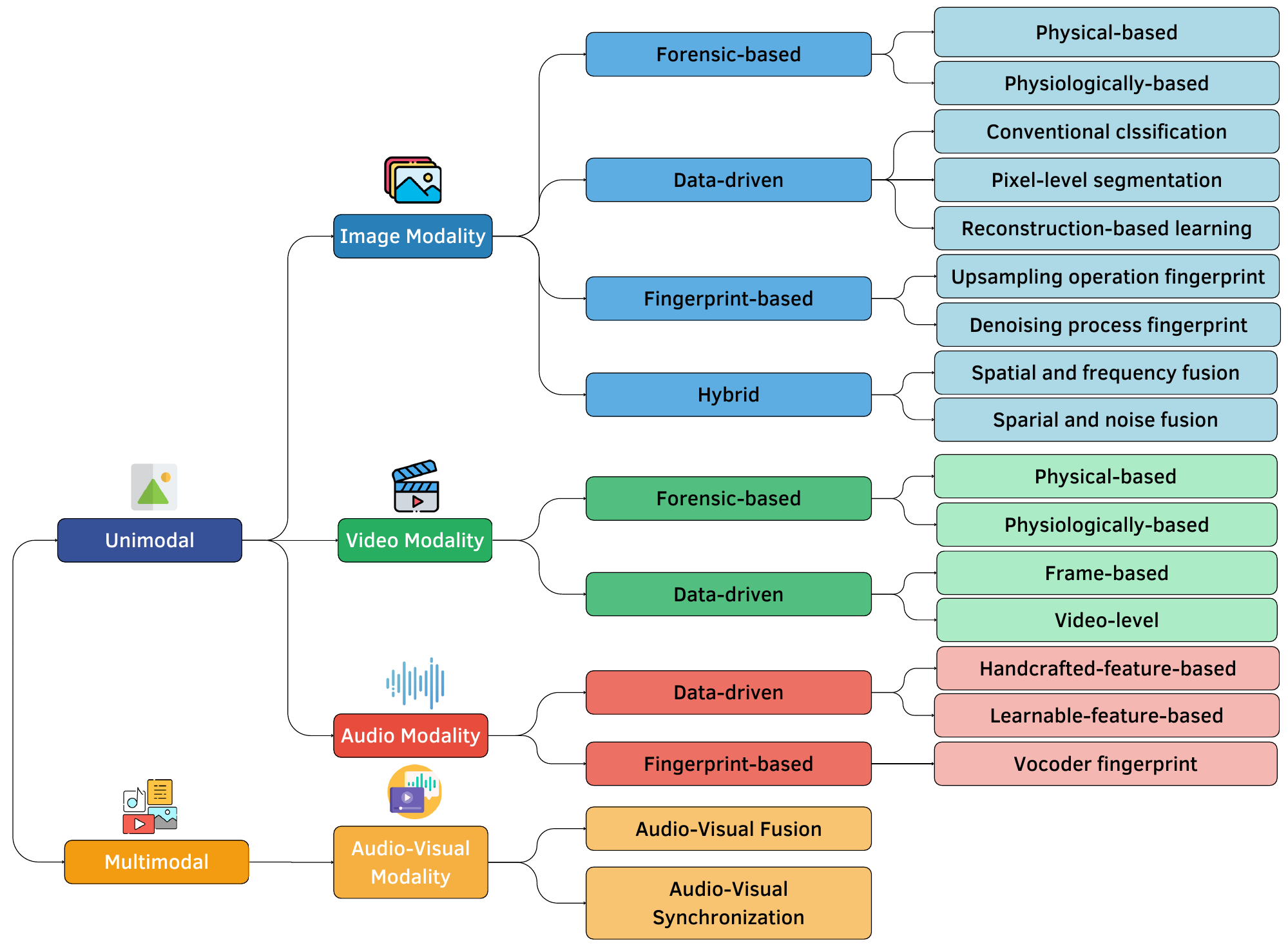}
	\caption{Taxonomy of passive DF detection approaches.}
	\label{fig:taxonomy}
	\vspace*{-0.7\baselineskip}
\end{figure}

\section{Unimodal Detection} \label{sec:unimodal}
This section provides a comprehensive overview of unimodal approaches that focus on analyzing a single modality, typically either the visual or audio component of media, to identify manipulated content. In the following subsections, we provide a comprehensive review of SoTA approaches for each modality, discussing their methodologies, strengths, and limitations. Table \ref{tab:passive-detection} summarizes key ideas, strengths, and limitations of unimodal and multimodal approaches.

\begin{table}[ht]
	\centering
	\fontsize{5pt}{5pt}\selectfont
	\caption{A Summarization of Ideas, Strengths, and Limitations of Approaches across Multi-Modality}
	\label{tab:passive-detection}
	\begin{tblr}{
			width = \linewidth,
			colspec = {Q[60]Q[70]Q[90]Q[170]Q[230]Q[270]},
			row{1} = {c},
	row{2} = {c},
	row{4} = {c},
	row{6} = {c},
	row{7} = {c},
	row{9} = {c},
	row{11} = {c},
	row{12} = {c},
	row{14} = {c},
	row{17} = {c},
	row{18} = {c},
	row{22} = {c},
	cell{2}{1} = {c=6}{0.936\linewidth},
	cell{3}{1} = {r=2}{c},
	cell{3}{2} = {c},
	cell{3}{3} = {c},
	cell{3}{4} = {c},
	cell{3}{5} = {r=2}{},
	cell{3}{6} = {r=2}{},
	cell{5}{1} = {r=3}{c},
	cell{5}{2} = {c},
	cell{5}{3} = {c},
	cell{5}{4} = {c},
	cell{5}{5} = {r=3}{},
	cell{5}{6} = {r=3}{},
	cell{8}{1} = {r=2}{c},
	cell{8}{2} = {c},
	cell{8}{3} = {c},
	cell{8}{4} = {c},
	cell{8}{5} = {r=2}{},
	cell{8}{6} = {r=2}{},
	cell{10}{1} = {r=2}{c},
	cell{10}{2} = {c},
	cell{10}{3} = {c},
	cell{10}{4} = {c},
	cell{10}{5} = {r=2}{},
	cell{10}{6} = {r=2}{},
	cell{12}{1} = {c=6}{0.936\linewidth},
	cell{13}{1} = {r=2}{c},
	cell{13}{2} = {c},
	cell{13}{3} = {c},
	cell{13}{4} = {c},
	cell{13}{5} = {r=2}{},
	cell{13}{6} = {r=2}{},
	cell{15}{1} = {r=3}{c},
	cell{15}{2} = {c},
	cell{15}{3} = {c},
	cell{15}{4} = {c},
	cell{16}{2} = {r=2}{c},
	cell{16}{3} = {c},
	cell{16}{4} = {c},
	cell{16}{5} = {r=2}{},
	cell{16}{6} = {r=2}{},
	cell{18}{1} = {c=6}{0.936\linewidth},
	cell{19}{1} = {r=2}{c},
	cell{19}{2} = {c},
	cell{19}{3} = {c},
	cell{19}{4} = {c},
	cell{20}{2} = {c},
	cell{20}{3} = {c},
	cell{20}{4} = {c},
	cell{21}{2} = {c},
	cell{21}{3} = {c},
	cell{21}{4} = {c},
	cell{22}{1} = {c=6}{0.936\linewidth},
	cell{23}{1} = {c=2}{0.115\linewidth,c},
	cell{23}{3} = {c},
	cell{23}{4} = {c},
	cell{24}{1} = {c=2}{0.115\linewidth,c},
	cell{24}{3} = {c},
	cell{24}{4} = {c},
	vline{2-6} = {1-24}{},
	hline{1,25} = {-}{0.08em},
	hline{2-3,5,8,12-13,15,18-19,21-24} = {-}{},
	hline{2} = {2}{-}{},
	hline{4,6-7,9,11,14} = {2-4}{},
	hline{10,16,20} = {2-6}{},
	hline{17} = {3-4}{},
	}
	\textbf{Categories}                   & \textbf{Approach}               & \textbf{Article}                                                                                                                                                                                                                                                                                                                                                                & \textbf{Key Idea}                                                                                                                         & \textbf{Strengths}                                                                                                                                                     & \textbf{Limitations}                                                                                                                                                                                                                             \\
	\textbf{Image Modality}               &                                 &                                                                                                                                                                                                                                                                                                                                                                                 &                                                                                                                                           &                                                                                                                                                                        &                                                                                                                                                                                                                                                  \\
	\textbf{Forensic-based}               & Physical-based                  & \cite{chen2021robust}, \cite{amin2024exploring},\cite{qiao2023csc}, \cite{talib2025chrominance}                                                                                                                                                                                                                                                                                 & Color statistics analysis                                                                                                                 & {- Provide strong interpretability for the detection decisions;\\- Don't rely on generator architectures or maniputation techniques}                                   & {- May become less effective for DM-based images that can preserve semantic content in images \cite{vahdati2024beyond};\\- Struggle to low-quality fake images or images applied postprocessing techniques \cite{barni2020cnn}}                  \\
										& Physiologically-based           & ~\cite{nirkin2021deepfake}, \cite{hu2021exposing}, \cite{wang2022eyes}                                                                                                                                                                                                                                                                                                          & Biological consistency between eyes or facial symmetry                                                                                    &                                                                                                                                                                        &                                                                                                                                                                                                                                                  \\
	\textbf{Data-driven}                  & Conventional classification     & \cite{huang2022fakelocator},\cite{wang2022lisiam}, \cite{mazaheri2022detection}, \cite{katamneni2024contextual}, \cite{das2022gca}, \cite{guo2023hierarchical}, \cite{tantaru2024weakly}                                                                                                                                                                                        & Cast as simple binary classification or fine-grained classification problem                                                               & {- Auctomatically~learn discriminative features from data;\\- End-to-end training}                                                                                     & {- Rely on the quality~and quantity of training data; \\- Limited generalization across~different datasets or manipulation techniques; \\- Lack of interpretability since detectors are black-box}                                               \\
										& Pixel-level segmentation        & \cite{frank2020leveraging}, \cite{qian2020thinking}, \cite{dzanic2020fourier}, \cite{liu2020global}, \cite{tan2024rethinking}, \cite{hong2024contrastive}                                                                                                                                                                                                                       & Cast as segmentation problem to predict manipulated regions                                                                               &                                                                                                                                                                        &                                                                                                                                                                                                                                                  \\
										& Reconstruction-based learning   & \cite{he2021beyond}, \cite{cao2022end}, \cite{liu2023fedforgery}, \cite{shi2023discrepancy}, \cite{guo2024deepfake}                                                                                                                                                                                                                                                             & Formulate through the lens of~image reconstruction                                                                                        &                                                                                                                                                                        &                                                                                                                                                                                                                                                  \\
	\textbf{Fingerprint-based}            & Upsamping operation fingerprint & \cite{frank2020leveraging}, \cite{qian2020thinking}, \cite{dzanic2020fourier}, \cite{liu2020global}, \cite{tan2024rethinking}                                                                                                                                                                                                                                                   & Upsampling operations leave checkerboard patterns in the frequency domain                                                                 & {- Less sensitive to image content since methods focus on model-specific artifacts; \\- Provide clearer reasoning for their decision compared to black-box DL methods} & {- Generators can remove these fingerprints through additional constraints \cite{chandrasegaran2021closer, neves2020ganprintr};\\- Vulnerable to postprocessing techniques which can remove these fingerprints}                                  \\
										& Denoising process fingerprint   & \cite{wang2023dire}, \cite{ma2023exposing}                                                                                                                                                                                                                                                                                                                                      & Denoising process leaves lower reconstruction errors between fake and reconstructed images                                                &                                                                                                                                                                        &                                                                                                                                                                                                                                                  \\
	\textbf{Hybrid}                       & Spatial and frequency fusion    & \cite{miao2023f}, \cite{miao2022hierarchical}, \cite{wang2022m2tr}, \cite{wang2023dynamic}                                                                                                                                                                                                                                                                                      & Leverage spatial and frequency domain                                                                                                     & {- Capture a broader range of artifacts through different modalities;\\- More robust against~single-modality evasion techniques}                                            & {- Processing multiple feature streams requires more computational resources and may increase inference time;\\-~Multi-stream architectures can be harder to train}                                                                              \\
										& Spatial and noise fusion        & \cite{kong2022detect}, \cite{kong2022detect}, \cite{shuai2023locate}, \cite{guillaro2023trufor}                                                                                                                                                                                                                                                                                 & Leverage RGB features and noise patterns                                                                                                  &                                                                                                                                                                        &                                                                                                                                                                                                                                                  \\
	\textbf{ Video Modality}              &                                 &                                                                                                                                                                                                                                                                                                                                                                                 &                                                                                                                                           &                                                                                                                                                                        &                                                                                                                                                                                                                                                  \\
	\textbf{Forensic-based}               & Physical-based                  & \cite{xia2022towards}, \cite{huda2024fake}                                                                                                                                                                                                                                                                                                                                      & Color statistics or surface inconsistencies analysis~ frame-by-frame                                                                      & {- Provide strong interpretability for the detection decisions;\\- Don't rely on generator architectures or maniputation techniques}                                   & {- May become less effective for DM-based images that can preserve semantic content in images \cite{vahdati2024beyond};\\- Struggle to low-quality fake images or images applied postprocessing techniques \cite{barni2020cnn}}                  \\
										& Physiologically-based           & \cite{sun2021improving}, \cite{li2023forensic}, \cite{jeon2022deepfake}                                                                                                                                                                                                                                                                                                         & Skin color changes, facial movements, and~symmetrical face patches                                                                        &                                                                                                                                                                        &                                                                                                                                                                                                                                                  \\
	\textbf{Data-driven}                  & Frame-based                     & \cite{wang2023noise}, \cite{ciamarra2024deepfake},~\cite{ba2024exposing}, \cite{bonettini2021video},~\cite{gu2021spatiotemporal}, \cite{ gu2022delving}, \cite{gu2022region}, \cite{gu2022hierarchical}                                                                                                                                                                         & Based on CNN architectures to classify each frame as real or fake and then fuse them for the final decision                               & - Take advantage of image-level approaches;                                                                                                                            & {- Lack temporal information to detect DFs;\\- Low performance when detecting DF videos \cite{vahdati2024beyond}.}                                                                                                                               \\
										& Video-level                     & \cite{montserrat2020deepfakes}, \cite{saikia2022hybrid}, \cite{coccomini2022combining}, \cite{zhang2022deepfake}, \cite{zhang2021detecting},~\cite{zheng2021exploring}                                                                                                                                                                                                          & Design architectures or techniques that can capture inter-frame inconsistencies                                                           & - Capture both spatial and temporal relationships to identify inconsistencies in video                                                                                 & {- High computational and GPU memory requirements for processing multiple frames simultanously;\\- Identity-driven approaches may struggle when dealing with face reenactment where the identity is preserved throughout the manipulated video.} \\
										&                                 & {\cite{agarwal2020detecting}, \cite{huang2023implicit},~\cite{cozzolino2021id},\\\cite{liu2023ti2net}, \cite{dong2022protecting}, \cite{coccomini2024mintime}}                                                                                                                                                                                                                  & Explore identity inconsistencies over the video                                                                                           &                                                                                                                                                                        &                                                                                                                                                                                                                                                  \\
	\textbf{Audio Modality}               &                                 &                                                                                                                                                                                                                                                                                                                                                                                 &                                                                                                                                           &                                                                                                                                                                        &                                                                                                                                                                                                                                                  \\
	\textbf{Data-driven}                  & Handcrafted-feature-based~      & \cite{alzantot2019deep}, \cite{yang2018extended}, \cite{xue2022audio}, \cite{kwak2021resmax}, \cite{gao2021detection}, \cite{saleem2019voice}, \cite{wani2024abc}                                                                                                                                                                                                               & Utilize expert-based audio processing techniques to extract physical and perceptual features~ for DF audio detection                      & {- Extract~predefined features typically requires less computational power;\\-~Make detection decisions more transparent and explainable.}                             & {- Lack of generalization of out-of-domains \cite{yang2024robust};\\-~Predefined features cannot adapt to the specific characteristics of different datasets}                                                                                    \\
										& Learnable-feature-based         & {\cite{jung2022aasist}, \cite{huang2023discriminative},~\cite{hua2021towards},~\\\cite{tak2021end},~\cite{shin2024hm},~\cite{chakravarty2024lightweight},~\\\cite{chen2024rawbmamba}, \cite{yang2024robust},~~\cite{wang2024can}, \cite{tran2024spoofed}, \cite{guo2024audio},~\cite{martin2024exploring},\\\cite{pan2024attentive},~\cite{doan2024balance},~\cite{kim2024one}} & Leverage supervised-trainable front-ends or pre-trained self-supervised models to optimal feature representations                         & {- Offer an E2E training process;\\- Capture sophisticated temporal patterns across multiple time scales.}                                                             & {- Use SSL models as front-end feature extraction can lead to computational overhead and overfitting to limited downstream data;\\- Learned features often lack clear acoustic meaning,~making their decisions harder to explain}                \\
	\textbf{Fingerprint-based}            & Vocoder fingerprint             & \cite{yan2022initial}, \cite{sun2023ai}                                                                                                                                                                                                                                                                                                                                         & Explore artifacts left by vovoders                                                                                                        & {- Provide better explanation for the decisions;\\- Less sensitive to speech content}                                                                                  & - Vulnerable to postprocessing techniques that can remove artifacts                                                                                                                                                                              \\
	\textbf{Multimodal Modality}          &                                 &                                                                                                                                                                                                                                                                                                                                                                                 &                                                                                                                                           &                                                                                                                                                                        &                                                                                                                                                                                                                                                  \\
	\textbf{ Audio-visual fusion}         &                                 & {\cite{ilyas2023avfakenet},~\cite{raza2023multimodal},~\cite{hashmi2022multimodal},\\\cite{wang2024audio}, \cite{zhang2024joint}, \cite{katamneni2023mis},\\\cite{wang2024avt},~\cite{zhou2021joint}, \cite{wang2024building}}                                                                                                                                                  & Combine features from both modalities to leverage complementary information through concatenation or more sophisticated fusion strategies & - Can leverage networks in visual and audio modalities to extract independent features                                                                                    & {- Difficult to determine the optimal fusion strategies;\\- Cannot capture temporal synchronization between modalities}                                                                                                                          \\
	\textbf{Audio-visual Synchronization} &                                 & {\cite{astrid2025audio},~\cite{shahzad2023av},~\cite{yang2023avoid},~\\\cite{chugh2020not}, \cite{liu2023mcl}, \cite{zhang2024joint},\\\cite{cheng2023voice}, \cite{oorloff2024avff}, \cite{yu2023pvass}, \cite{haliassos2022leveraging},~\cite{feng2023self}}                                                                                                                  & Focus on modeling the temporal relationship between speech and facial movements                                                           & - Can detect subtle audio-visual inconsistencies~ that fusion methods can miss                                                                                         & - Sensitive to natural variations (e.g., speaking styles) or natural~asynchrony (e.g.,~errors in encoding orrecording)
	\end{tblr}
	\vspace*{-2.5\baselineskip}
\end{table}

\subsection{Image Modality}
\begin{figure*}[!ht]
	\centering
	\includegraphics[width=0.99\linewidth]{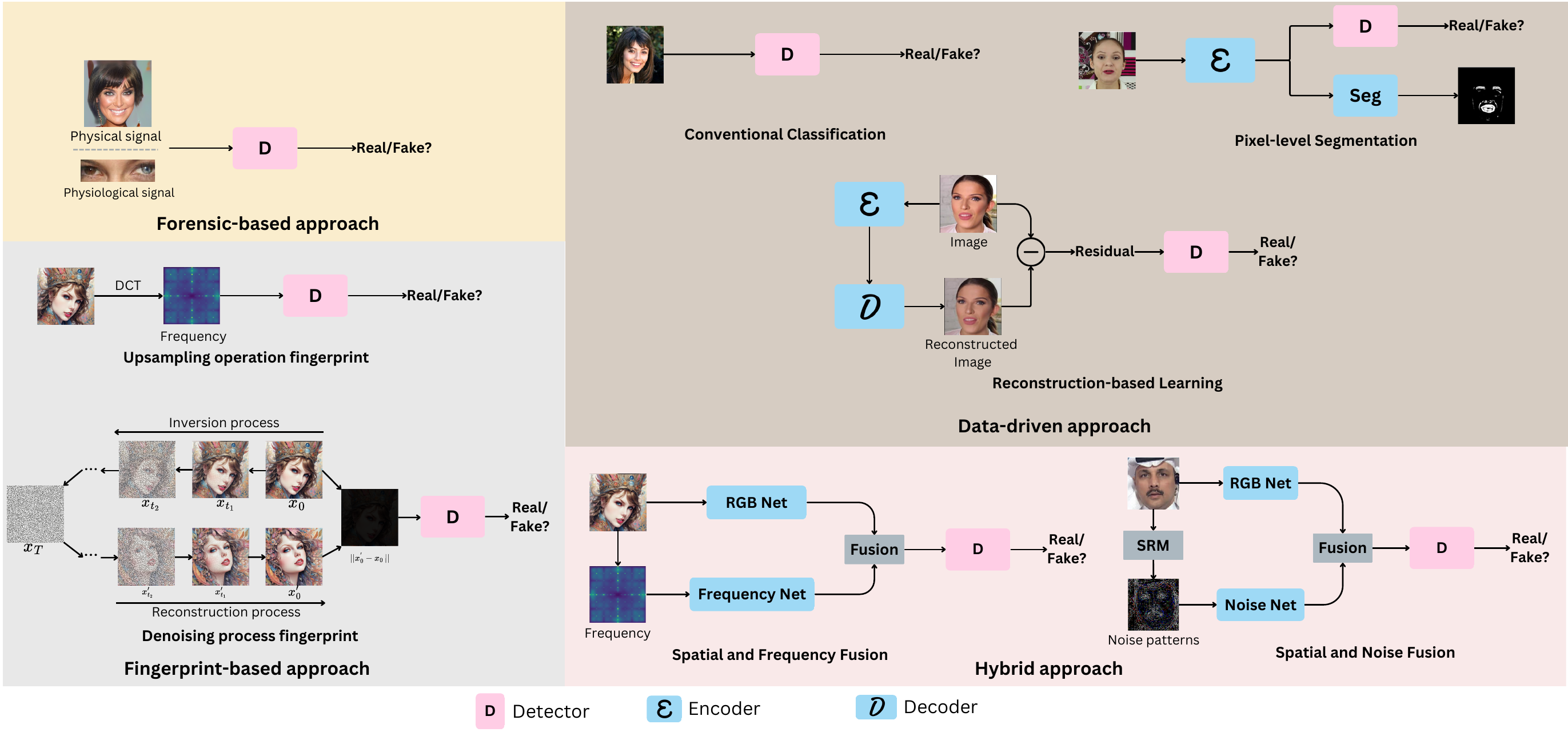}
	\caption{Illustration of passive DF detection approaches in image modality.}
	\label{fig:image-domain}
	\vspace*{-0.7\baselineskip}
\end{figure*}

We divide the methods in the image modality into four main categories: Forensic-based, Data-driven, Fingerprint-based, and Hybrid. Figure \ref{fig:image-domain} illustrates four main categories in the image modality.

\textbf{Forensic-based methods.}
Methods in this group detect DFs based on predefined rules related to differences in physical-based and physiologically-based signals between real and fake images. For example, the inconsistencies between left and right eyes or the unnaturalness of lighting conditions in fake images. The key insight behind these approaches is that generators cannot perfectly replicate the complex physical properties and biological consistencies inherent in natural image formation. Some studies reveal that natural images exhibit consistent relationships between luminance and chrominance components \cite{chen2021robust, talib2025chrominance}; thus, analyzing multiple color spaces can provide a more comprehensive view of potential manipulation traces. Other works use cross-color spatial co-occurrence matrices \cite{qiao2023csc} or Wasserstein distance \cite{amin2024exploring} to analyze statistical relationships between color channels of real and fake images. Biological-based inconsistencies are also analyzed to identify DFs, such as the differences between left and right eyes  \cite{hu2021exposing,wang2022eyes} and the discrepancies between the internal face region and the surrounding context \cite{nirkin2021deepfake}.

\textbf{Data-driven methods.} Instead of relying on predefined rules, these methods leverage deep learning (DL) to automatically learn discriminative features from large-scale datasets. Based on how they formulate the detection problem, data-driven approaches can be categorized into three main groups.

\circled{1} \textit{Conventional classification} methods cast DF detection as a binary classification problem, where the model learns to classify an input image as real or fake \cite{yang2021mtd,sha2023fake}. However, simply treating DF detection as a binary classification problem may not be optimal due to the subtle and localized nature of the differences between real and fake images \cite{zhao2021multi, han2023fcd}. \citeauthor{zhao2021multi} \cite{zhao2021multi} reformulated as a fine-grained classification task that enables the detector to focus on local regions, while \citeauthor{han2023fcd} \cite{han2023fcd} reformulated as multi-task learning by designing a network that can detect multiple types of DF face images simultaneously.

\circled{2} \textit{Pixel-level segmentation} methods formulate detection as a dense prediction task that generates localization maps to identify manipulated regions at the pixel level. By predicting manipulation masks, these methods not only determine the input as real or fake but also pinpoint which image regions have been manipulated. This additional localization capability provides more interpretable results and can help understand the manipulation techniques used. These methods localize the manipulated area in a fully supervised setting through an encoder-decoder architecture \cite{huang2022fakelocator, wang2022lisiam, mazaheri2022detection, katamneni2024contextual} or attention mechanism \cite{das2022gca, guo2023hierarchical, hong2024contrastive}. However, these require dense pixel-wise ground-truth masks, which might not always be available. \citeauthor{tantaru2024weakly} \cite{tantaru2024weakly} overcomes this limitation by applying the GradCAM explainability technique to the activations produced by a classification network to highlight the regions most predictive of the fake class.

\circled{3} \textit{Reconstruction-based learning} methods formulate detection through the lens of image reconstruction. These approaches typically attempt to reconstruct the input image using an encoder-decoder architecture and analyze the reconstruction errors. The key principle is to focus on learning what real images should look like through reconstruction, and the reconstruction of fake images is different from that of real images. However, \citeauthor{shi2023discrepancy} \cite{shi2023discrepancy} recognized that single reconstruction-based methods \cite{he2021beyond, cao2022end, liu2023fedforgery, guo2024deepfake} have limited feature representation and proposed a double-head reconstruction module that combines discrepancy-guided encoding, dual reconstruction, and aggregation-based detection to improve forgery detection performance.

\textbf{Fingerprint-based methods.} Fingerprint-based methods for DF detection explore the unique patterns or artifacts unintentionally embedded into the generated images by the architectural design and training process of GMs. The key insight is that different GMs leave distinct \textit{fingerprints} in their outputs due to their architectural design and training process. For GAN-generated images, these methods focus on artifacts introduced by the upsampling operations commonly used in generator architectures. Several studies have shown that the frequency spectrum of GAN-generated images often contains distinctive patterns - "checkerboards", particularly in the middle and high-frequency bands \cite{frank2020leveraging, qian2020thinking, dzanic2020fourier, liu2020global, tan2024rethinking}. Regarding images generated by DMs, the iterative denoising process of DMs tends to leave characteristic traces that can be detected through careful analysis. For instance, some methods measure the reconstruction error when passing an image through a pre-trained DM \cite{wang2023dire, ma2023exposing} based on the observation that synthetic images can be more accurately reconstructed compared to real ones. Recently, \citeauthor{tan2024rethinking} \cite{tan2024rethinking} explores how upsampling operations in GANs and DMs create distinctive patterns in how neighboring pixels relate to each other. Rather than analyzing frequency-domain artifacts across the entire image like previous approaches, the authors propose to analyze the local interdependence between pixels in small 2x2 grid cells to capture these artifacts.

\textbf{Hybrid methods.} Detection approaches that solely rely on spatial domain information have shown to be highly susceptible to variations in dataset quality and post-processing operations \cite{qian2020thinking, wang2020cnn}. To address this limitation, hybrid approaches aim to leverage complementary information from different domains to enhance detection performance. The key insight is that DFs often leave traces across multiple feature spaces, and combining these cues can provide more robust detection than relying on a single domain. Based on the types of features being fused, hybrid methods can be categorized into two main groups.

\circled{1} \textit{Spatial and frequency fusion} methods combine features from the spatial domain (capturing visual content and local artifacts) with frequency domain information (revealing generation artifacts in the spectral space). The spatial stream typically employs convolutional networks to extract content-level features, while the frequency stream analyzes spectral patterns through discrete cosine transform (DCT) or discrete Fourier transform (DFT) \cite{miao2023f, miao2022hierarchical, wang2022m2tr, wang2023dynamic}. This dual-stream architecture helps capture both semantic inconsistencies and subtle frequency-domain artifacts introduced during the generation process.

\circled{2} \textit{Spatial and noise fusion} methods integrate standard RGB features with noise patterns extracted through SRM noise filters \cite{fridrich2012rich}. These approaches recognize that deepfake generation often leaves distinctive noise patterns that are different from those found in authentic images. By analyzing both content features and noise residuals, these methods can better distinguish between natural imaging noise patterns and synthetic artifacts \cite{kong2022detect}, \cite{kong2022detect}, \cite{shuai2023locate}. Instead of relying on handcrafted SRM noise filters, TruFor \cite{guillaro2023trufor} trains Noiseprint++ extractor in a self-supervised manner using contrastive learning to capture stronger noise traces.

\begin{figure}[ht]
	\centering
	\includegraphics[width=0.99\linewidth]{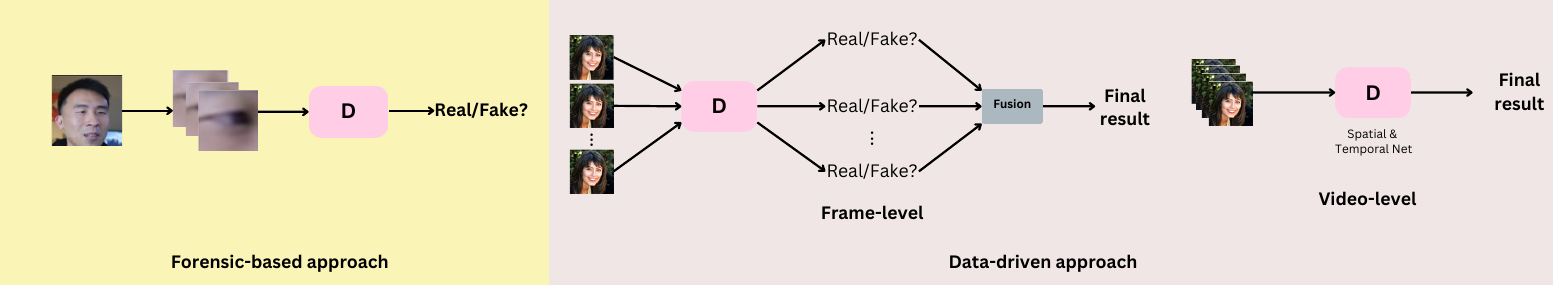}
	\caption{Illustration of passive DF detection approaches in video modality.}
	\label{fig:video-domain}
	\vspace*{-0.7\baselineskip}
\end{figure}

\subsection{Video Modality.} \hfill \break
\textbf{Compare with image modality}. Video-based DF detection presents unique challenges compared to image-based detection \cite{vahdati2024beyond}. On the one hand, approaches developed for image detection can be naturally extended to videos by applying them to individual frames. Forensic-based methods can analyze physical or physiological inconsistencies in each frame, while data-driven approaches can classify frames independently. The final decision can then be obtained by aggregating frame-level predictions through voting or averaging mechanisms. However, this frame-by-frame analysis fails to capture a crucial aspect of DF videos: temporal coherence. Manipulated videos often exhibit subtle temporal inconsistencies, such as unnatural head movements, inconsistent facial expressions across frames, or flickering artifacts in manipulated regions. These temporal artifacts cannot be detected by examining frames in isolation, making pure image-based approaches insufficient for robust video DF detection. Therefore, it is imperative to develop dedicated video-level approaches that explicitly model temporal relationships. In this survey, we categorize detection approaches in the video modality into two primary groups: (i) forensic-based and (ii) data-driven, which are illustrated in Figure \ref{fig:video-domain}.

\textbf{Forensic-based methods} primarily extend insights from image-modality forensic analysis to the video modality by examining physical and biological inconsistencies frame by frame. \citeauthor{xia2022towards} \cite{xia2022towards} analyzes multiple color channels, while \citeauthor{huda2024fake} \cite{huda2024fake} uses multiple texture feature descriptors to capture surface inconsistencies per frame. Other studies analyze physiological signals that should remain consistent in authentic videos but may be imperfectly replicated in DFs, such as blood flow patterns through skin color changes \cite{jeon2022deepfake}, facial movements \cite{sun2021improving, li2023forensic}, or the difference between front and side face images \cite{li2023forensic}. Advanced methods \cite{sun2021improving, li2023forensic} utilize Recurrent Neural Network (RNN) to model the temporal characteristics on the embedded feature sequences.

\textbf{Data-driven approaches.} Based on how temporal information is processed, these methods can be categorized into two main approaches: Frame-level and Video-level.

\circled{1} \textit{Frame-level methods} essentially apply image-modality DL techniques to individual frames, treating the video as a sequence of independent images. These methods are typically based on CNN architectures to classify each frame as real or fake, then aggregate these frame-level predictions through techniques like majority voting or temporal averaging to make a final video-level decision \cite{wang2023noise, ciamarra2024deepfake, bonettini2021video}. To capture a broad set of forgery clues (e.g., blending ghosts, skin tone inconsistencies, tooth details, stitching seams), \citeauthor{ba2024exposing} \cite{ba2024exposing} proposes a local disentanglement module to extract multiple local representations and then fuses them into a global semantic-rich feature. Rather than relying on per-frame, some works \cite{gu2021spatiotemporal, gu2022delving, gu2022region, gu2022hierarchical} have focused on capturing local inconsistency within densely sampled video snippets, where the entire video is densely divided into multiple snippets. While straightforward to implement, these frame-based approaches may miss temporal inconsistencies that are only visible when analyzing frame sequences.

\circled{2} \textit{Video-level methods} explicitly model temporal relationships by utilizing architectures designed to capture both spatial and temporal information. Early attempts combined convolution-based networks and recurrent-based networks to extract global spatio-temporal features \cite{montserrat2020deepfakes, saikia2022hybrid}; however, this approach has been demonstrated to be less effective \cite{zhao2023istvt}. Recent approaches have utilized advanced architectures or techniques that can capture long-range dependencies and inter-frame inconsistencies, such as transformers \cite{zhao2023istvt, coccomini2022combining, zhao2022self, zhang2022deepfake}, 3D CNN \cite{zhang2021detecting}, temporal convolution networks (TCN) \cite{zheng2021exploring} and attention mechanism \cite{hu2024delocate}.
%
Another line of work aims at identifying the identity-related inconsistencies in DF videos. These approaches recognize that while DFs may maintain visual quality in individual frames, they often struggle to consistently preserve identity characteristics across an entire video sequence. These methods typically employ two main strategies. The first strategy is to utilize pre-trained face recognition to extract identity embeddings and analyze how these embeddings evolve over time to detect unnatural variations \cite{agarwal2020detecting, huang2023implicit}. The second strategy focuses on capturing spatio-temporal identity features directly through end-to-end training through 3D face reconstruction or identity-specific transformer  \cite{cozzolino2021id,liu2023ti2net,dong2022protecting}. Recently, MinTime \cite{coccomini2024mintime} aims to address a crucial limitation in previous identity-driven approaches that struggle with videos containing multiple people. Particularly, the authors introduce an identity-aware attention mechanism that applies masking to process each identity in the video independently, enabling the detector to handle multiple identities in the video.

\begin{figure}[!ht]
	\centering
	\includegraphics[width=0.99\linewidth]{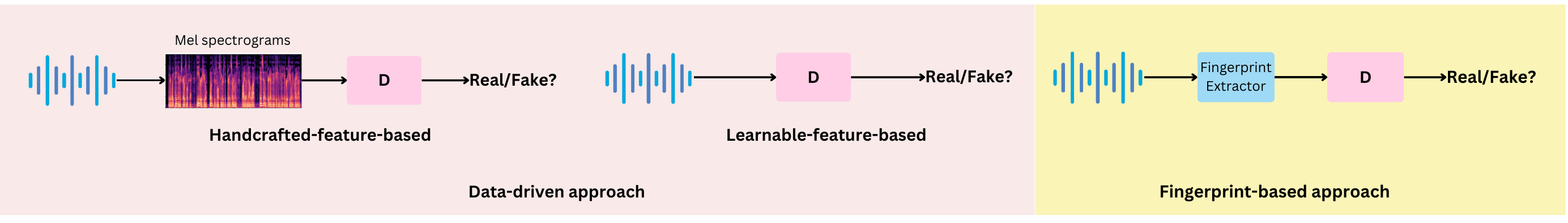}
	\caption{Illustration of passive DF detection approaches in audio modality.}
	\label{fig:audio-domain}
	\vspace*{-0.7\baselineskip}
\end{figure}

\subsection{Audio Modality.} \hfill \break
\textbf{Compare with image modality.} Methodologies developed for visual media cannot be directly applied to detect DF audio \cite{lee2024tug}. Image DF detection typically relies on visual artifacts such as inconsistent lighting, unnatural eye reflections, or imperfect facial blending, which can be analyzed using CNNs trained on pixel-level data. In contrast, audio signals exist primarily as 1D temporal waveforms or 2D time-frequency representations where artifacts manifest as spectral anomalies, inconsistent phase coherence, or irregular prosodic patterns. For example, synthetic voices often exhibit unnatural pauses, overly consistent pitch modulation, or artifacts in high-frequency bands that are imperceptible to humans but detectable via mel-spectrogram analysis. Given this distinct nature of audio signals, current \textit{data-driven approaches} differ primarily in extracting discriminative acoustic features before classification. Current data-driven approaches typically follow a two-stage pipeline: a front-end for feature extraction and a back-end for classification. The front-end transforms raw audio signals into acoustic representations, while the back-end analyzes these features to make a binary real/fake decision. Furthermore, another group of audio DF detection - \textit{fingerprint-based approaches} that explore artifacts left by speech synthesis models, especially the vocoders. Figure \ref{fig:audio-domain} illustrates these approaches.

\textbf{Data-driven methods.} Based on the front-end feature extraction techniques employed, data-driven methods can be classified into two main categories.

\circled{1} \textit{Handcrafted-feature-based methods} utilize expert-based audio processing techniques as front-end to convert the audio waveform into predefined acoustic features. These features fall into two main categories - physical and perceptual - each capturing different aspects that help distinguish between authentic and synthetic speech \cite{li2022comparative}. Mel Frequency Cepstral Coefficients (MFCC) capture the spectral envelope of speech, which often exhibits unnatural patterns in synthetic audio due to imperfect modeling of vocal tract characteristics \cite{alzantot2019deep}. Constant Q transform (CQT) and spectrogram representations reveal frequency distributions and power spectrum patterns that may be distorted in DF audio \cite{yang2018extended, xue2022audio, kwak2021resmax, wani2024abc}.
In contrast, perceptual features are designed to capture characteristics that align with human auditory perception and natural speech properties. Techniques like Jitter and Shimmer measure frequency and amplitude variations, respectively, capturing micro-perturbations that are naturally present in human voice but often missing or incorrectly reproduced in synthetic speech \cite{gao2021detection}. Chromagram analysis examines pitch-related features, helping identify unnatural pitch progressions or tonal qualities that may occur in DF audio \cite{saleem2019voice}. These extracted features provide rich information for the classifier to distinguish between real and fake audios.

\circled{2} \textit{Learnable feature-based methods} leverage NNs to automatically extract discriminative features directly from raw audio in an end-to-end (E2E) manner. These methods design specialized network architectures that learn optimal feature representations through supervised learning, such as the spectro-temporal graph attention network \cite{jung2022aasist, huang2023discriminative}, Inception-style architecture \cite{hua2021towards}, RawNet-based architecture \cite{tak2021end}, and Conformer-based architecture \cite{shin2024hm}. Taking the advantage of Mamba \cite{gu2023mamba}, \citeauthor{chen2024rawbmamba} \cite{chen2024rawbmamba} proposes a novel E2E bidirectional state space model for audio DF detection to capture more long-range relationships. \citeauthor{yang2024robust} \cite{yang2024robust} introduces a multi-view feature fusion approach that concatenates handcrafted and learnable features to capture a broad range of audio features. For resource-constrained applications, \citeauthor{chakravarty2024lightweight} \cite{chakravarty2024lightweight} propose a lightweight feature extractor using linear discriminant analysis to reduce a 2048-dimensional feature vector to a single, highly discriminative feature.

Inspired by advancements in speech self-supervised learning, several works \cite{wang2024can,tran2024spoofed,guo2024audio} leverage self-supervised learning (SSL) speech models like Wav2vec2 and XLS-R as front-end feature extractors for DF detection. These models, pre-trained on amounts of unlabeled speech data, demonstrate remarkable capacity for capturing subtle acoustic characteristics across diverse speaking conditions. However, naively fine-tuning these large pre-trained models for DF detection presents challenges, including overfitting with limited training data and computational overhead \cite{martin2024exploring,wu2024adapter}. To address these concerns, \citeauthor{martin2024exploring} \cite{martin2024exploring} a lightweight downstream classifier with minimal trainable parameters to preserve the generalized audio representations of the SSL model while \citeauthor{pan2024attentive} \cite{pan2024attentive} selectively fine-tunes only early and middle layers to reduce computational requirements. Beyond traditional supervised learning paradigms, some researchers have explored alternative training strategies, including re-synthesizing real samples using various neural vocoders \cite{doan2024balance} or modeling only genuine speech representations and classifying any sample falling outside these established boundaries as fake \cite{kim2024one}.

\textbf{Fingerprint-based methods.} Similar to the image modality, these methods recognize that different vocoder architectures produce characteristic artifacts in the frequency spectrum, phase patterns, or temporal structure of the generated audio. The core idea is to identify and extract these vocoder-specific "fingerprints" that may be detectable through signal analysis. For instance, neural vocoders often struggle to reproduce certain aspects of natural speech, such as phase coherence across frequency bands or precise harmonic structures. Particularly, these methods have explored that the spectrograms of original audio have more natural and consistent high-frequency components and more natural harmonic structures compared to synthetic audio \cite{yan2022initial, sun2023ai}.

\begin{figure}[!ht]
	\centering
	\includegraphics[width=0.99\linewidth]{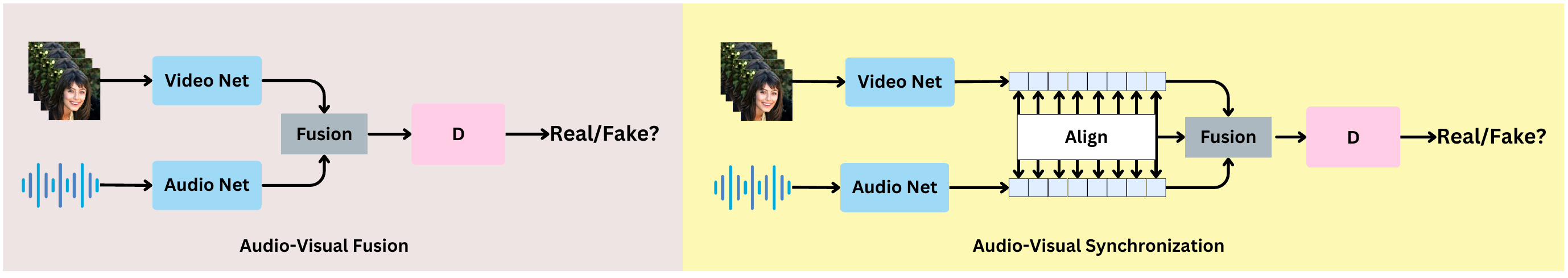}
	\caption{Illustration of passive DF detection approaches in multimodal modality.}
	\label{fig:multimodal-domain}
\end{figure}

\section{Multimodal Detection} \hfill \break
\label{sec:multimodal}
\textbf{Motivation for multimodal DF detection.} The rapid advancement of GMs capable of simultaneously manipulating both visual and audio components has necessitated the development of multimodal detection approaches. Recent DF generation has evolved beyond single-modality manipulation to creating talking-face videos where both face movements and voice are synthetically generated \cite{xu2024vasa, aneja2024facetalk, kim2024all, zhang2023sadtalker}. These advanced DFs maintain convincing audio-visual consistency, presenting significant challenges for unimodal detection systems. Visual-only detectors cannot identify audio manipulations, while audio-only detectors remain blind to visual forgeries \cite{yang2023avoid, feng2023self}. Addressing this critical gap requires multimodal approaches that leverage the intrinsic temporal and semantic correlations present in authentic human communication. Genuine video faces exhibit an intrinsic correlation between the mouth articulations and the speech units and an alignment of emotional nuances embedded in the facial and speech expressions \cite{oorloff2024avff}. These biologically-constrained relationships are challenging for DF generators to replicate perfectly. By analyzing the cross-modal relationships between facial movements and speech patterns, multimodal detection systems can identify subtle inconsistencies that remain imperceptible when examining individual modalities in isolation.

Building on this motivation, current multimodal DF detection approaches can be broadly categorized into two main groups based on their underlying detection principles: (i) \textit{Audio-visual fusion methods} which combine features from both modalities to leverage complementary information, (ii) \textit{Audio-Visual Synchronization methods} which focus on temporal alignment and semantic coherence between speech and facial movements. Figure \ref{fig:multimodal-domain} shows the difference between two approaches.

\subsection{Audio-visual fusion methods} These approaches typically employ dual-stream architectures where separate networks process audio and visual inputs, with their features later integrated through concatenation or more sophisticated fusion strategies. AVFakeNet \cite{ilyas2023avfakenet} basically relies on independent detection in each modality and combines predictions per modality for the final decision, while the works \cite{raza2023multimodal, hashmi2022multimodal} concats these features into a single vector before feeding them into a classifier. Cross-modal attention mechanisms are implemented in \cite{wang2024audio, zhang2024joint, katamneni2023mis, wang2024building} to align and model the interactions between audio and visual features. Other works design sophisticated fusion schemes to better capture the intrinsic audio-visual patterns, such as features fusion across multiple intermediate layers \cite{zhou2021joint} and dynamically weights the importance of each modality \cite{wang2024avt}. Although audio-visual fusion approaches can leverage SoTA networks in both visual and audio modalities to extract separate features, determining optimal fusion strategies remains challenging, as different integration methods may perform inconsistently across various datasets or manipulation techniques. Furthermore, most fusion approaches focus on feature integration rather than explicitly modeling the temporal synchronization between modalities, potentially missing subtle desynchronization artifacts.

\subsection{Audio-visual synchronization methods.} In contrast, these methods focus specifically on intrinsic temporal alignment and semantic coherence between speech and facial movements by directly modeling the relationship between them. These methods leverage the observation that even advanced DFs often struggle to maintain perfect cross-modal synchronization (e.g., precise lip movements during phoneme production or natural micro-delays between facial expressions and corresponding vocal elements). \citeauthor{astrid2025audio} \cite{astrid2025audio} captures fine-grained temporal inconsistencies by measuring audio-visual feature distances at each temporal step. \cite{shahzad2023av} and \cite{yang2023avoid} utilize multi-scale temporal convolutional network and vision transformer to capture inter-modal and intra-modal temporal correlations between audio-visual features. \textit{Contrastive learning (CL)} is also popularly utilized to build cross-modal correlations by making representations of corresponding audio-visual pairs similar while pushing non-corresponding pairs apart \cite{chugh2020not,liu2023mcl,zhang2024joint}. Another line of work proposes a two-stage framework: (i) representation learning and (ii) DF downstream classification \cite{cheng2023voice, oorloff2024avff, yu2023pvass, haliassos2022leveraging}. The first stage aims to acquire an audio-visual representation from real videos via CL to learn the dependency between real speech audio and the
corresponding visual facial features. In the second stage, this representation is applied for the DF detection task with a supervised learning objective to distinguish between real and fake videos. To capture better audio-visual correspondences, AVFF \cite{oorloff2024avff} introduces a masking strategy that forces the model to learn to predict masked content of one modality using information from the other modality. In contrast, \citeauthor{feng2023self} \cite{feng2023self} treats DF detection as an anomaly detection problem by using an autoregressive model to learn the normal distribution of synchronization features from real videos and flagging low-probability sequences as potential fakes.

\begin{table}[ht]
	\centering
	\fontsize{5pt}{5pt}\selectfont
	\caption{A Summarization of Generalization Methods}
	\label{tab:generalization}
	\begin{tblr}{
			width = \linewidth,
			colspec = {Q[75]Q[60]Q[250]Q[50]Q[45]Q[100]Q[45]Q[45]},
			cells = {c},
			cell{1}{1} = {r=2}{},
			cell{1}{2} = {r=2}{},
			cell{1}{3} = {r=2}{},
			cell{1}{4} = {r=2}{},
			cell{1}{5} = {c=4}{0.387\linewidth},
			cell{3}{1} = {r=11}{},
			cell{14}{1} = {r=5}{},
			cell{19}{1} = {r=4}{},
			cell{23}{1} = {r=5}{},
			cell{28}{1} = {r=3}{},
			cell{31}{1} = {r=9}{},
			vline{2-5} = {1-39}{},
			hline{1,40} = {-}{0.08em},
			hline{2} = {5-8}{},
			hline{3,14,19,23,28,31} = {-}{},
			hline{3} = {2}{-}{},
		}
		\textbf{Approach }         & \textbf{Article }                                             & \textbf{Key Idea }                                               & \textbf{Training} & \textbf{Evaluation$^{**}$} &                                                                  &             &              \\
		&                                                               &                                                                  &                   & \textbf{WD}                & \textbf{CD}                                                      & \textbf{CM} & \textbf{UD}  \\
		Data Augmentation          & \cite{ni2022core}                                             & Static image augmentation                                        & \circled{2}       & \circled{2}                & \circled{6},~\circled{7}                                         & -           & -            \\
		& \cite{luo2023beyond}                                          & Static image augmentation                                        & \circled{2}       & \circled{2}                & \circled{6},~\circled{7},~\circled{8},~\circled{9}               & \circled{1} & -            \\
		& \cite{doan2024balance}                                        & Static audio augmentation                                        & \circled{1}       & \circled{1}                & \circled{3}                                                      & -           & -            \\
		& \cite{martin2024exploring}                                    & Static audio augmentation                                        & \circled{1}       & \circled{1}                & \circled{3},~\circled{4},~\circled{5}                            & -           & -            \\
		& \cite{wang2023altfreezing}                                    & Static video augmentation                                        & \circled{2}       & \circled{2}                & ~\circled{7}, \circled{9}                                        & \circled{1} & -            \\
		& \cite{das2021towards}                                         & Dynamic image augmentation                                       & \circled{2}       & \circled{2}                & \circled{10},~\circled{7}                                        & -           & -            \\
		& \cite{wang2021representative}                                 & Dynamic image augmentation                                       & \circled{2}       & \circled{2}                & \circled{10},~\circled{7}                                        & -           & -            \\
		& \cite{yan2023transcending}                                    & Image-latent space augmentation                                  & \circled{2}       & \circled{2}                & \circled{7},~\circled{10}                                        & -           & -            \\
		& \cite{huang2025generalizable}                                 & Audio-latent space augmentation                                  & \circled{1}       & \circled{1}                & \circled{3}, \circled{5}                                         & -           & -            \\
		& \cite{xu2022supervised}                                       & Contrastive learning                                             & \circled{2}       & \circled{2}                & -                                                                & \circled{2} & -            \\
		& \cite{sun2022dual}                                            & Dual-contrastive learning                                        & \circled{2}       & \circled{2}                & \circled{6},~\circled{7},~\circled{8}                            & \circled{2} & -            \\
		Synthetic Data Training    & \cite{chen2022self}                                           & GAN framework + self-supervised auxiliary tasks                  & \circled{2}       & \circled{2}                & \circled{7},~\circled{6},~\circled{9}                            & -           & -            \\
		& \cite{shiohara2022detecting}                                  & Self-blending technique                                          & \circled{2}       & \circled{2}                & \circled{6},~\circled{7}                                         & \circled{2} &              \\
		& \cite{hasanaath2025fsbi}                                      & Self-blending frequency technique                                & \circled{2}       & \circled{2}                & \circled{7}                                                      & -           & -            \\
		& \cite{hanzhefreqblender}                                      & Self-blending in frequency domain                                & \circled{2}       & \circled{2}                & \circled{6},~\circled{7},~\circled{11}                           & \circled{2} & -            \\
		& \cite{yan2024generalizing}                                    & Video-level self-blending technique                              & \circled{2}       & \circled{2}                & \circled{6}, \circled{7},~\circled{8},~\circled{9}, \circled{11} & -           & \circled{12} \\
		Disentanglement Learning   & \cite{liang2022exploring}                                     & Dual-encoder architecture                                        & \circled{2}       & \circled{2}                & \circled{7}                                                      & \circled{2} & -            \\
		& \cite{yan2023ucf}                                             & Conditional decoder + contrastive learning                       & \circled{2}       & \circled{2}                & \circled{6},~\circled{7}                                         & \circled{2} & -            \\
		& \cite{guo2023controllable}                                    & Modle a controllable geometric embedding space                   & \circled{2}       & \circled{2}                & \circled{6},~\circled{7}                                         & -           & -            \\
		& \cite{chen2025diffusionfake}                                  & Identity desentanglement by reversing the generaive process      & \circled{2}       & \circled{2}                & \circled{6},~\circled{7},~\circled{8},~\circled{15}              & \circled{2} & -            \\
		Unsupervised Learning      & \cite{zhuang2022uia}                                          & Multivariate Gaussian estimation                                 & \circled{2}       & \circled{2}                & \circled{6},~\circled{7}                                         & \circled{2} & -            \\
		& \cite{qiao2024fully}                                          & Unsupervised contrastive learning                                & \circled{2}       & \circled{2}                & \circled{6}, \circled{7},~\circled{14}                           & -           & -            \\
		& \cite{haliassos2022leveraging}                                & transfer learning from self-supervised representations           & \circled{2}       & \circled{2}                & \circled{6},~\circled{7},~\circled{9}                            & -           & -            \\
		& \cite{oorloff2024avff}                                        & Transfer learning from self-supervised representations           & \circled{15}      & \circled{15}               & \circled{17},~\circled{6},~\circled{16}                          & -           & -            \\
		& \cite{kim2024one}                                             & Transfer learning from self-supervised representations           & \circled{1}       & \circled{1}                & \circled{3}                                                      & -           & -            \\
		Specific Training Strategy & \cite{choi2024exploiting}                                     & Two-stage training to learn style representation                 & \circled{2}       & \circled{2}                & \circled{7},~\circled{9}                                         & \circled{2} & -            \\
		& \cite{zhu2025slim}                                            & Two-stage training to learn style and linguistic representations & \circled{1}       & \circled{1}                & \circled{3},~\circled{5},~\circled{18}                           & -           & -            \\
		& \cite{cheng2024can}                                           & Progreassive training                                            & \circled{2}       & \circled{2}                & \circled{6}, \circled{7},~\circled{9},~\circled{12}              & -           & -            \\
		Adaptive Learning          & \cite{sun2021domain},~\cite{han2023sigma},~\cite{zhu2024tmfd} & Meta learning                                                    & \circled{2}       & \circled{2}                & \circled{6}, \circled{7}                                         &             &              \\
		& \cite{chen2022ost}                                            & Meta-learning + test-time training                               & \circled{2}       & \circled{2}                & \circled{6}, \circled{7},~\circled{9}                            & \circled{2} & -            \\
		& \cite{pan2023dfil}                                            & Incremental learning + knowledge distillation                    & \circled{2}       & \circled{2}                & \circled{6}, \circled{7}                                         & -           & -            \\
		& \cite{xie2023domain}                                          & Source domains aggredation + triplet loss                        & \circled{1}       & \circled{1}                & \circled{4},~\circled{15}                                        & -           & -            \\
		& \cite{lu2024one}                                              & One-class knowledge distillation                                 & \circled{1}       & \circled{1}                & \circled{3},~\circled{5}                                         & -           & -            \\
		& \cite{seraj2024semi}                                          & Domain alignment loss on source and target domain                & \circled{2}       & \circled{2}                & -                                                                & \circled{2} & -            \\
		& \cite{zhang2023adaptive}                                      & Low-rank adaptation matrices                                     & \circled{1}       & \circled{1}                & \circled{5}                                                      & -           & -            \\
		& \cite{wu2024adapter}                                          & Low-rank adaptation matrices                                     & \circled{1}       & \circled{1}                & \circled{5},~\circled{19}                                        & -           & -            \\
		& \cite{oiso2024prompt}                                         & Prompt tuning                                                    & \circled{1}       & \circled{1}                & \circled{3},~\circled{5}                                         & -           & -
	\end{tblr}
	\begin{tablenotes}
		\item i) \textbf{Training}: The dataset that the method trained on
		\item ii) \textbf{Evaluation}: Protocols that the method evaluated on: Within-Domain (WD), Cross-Domain (CR), Cross-Manipulation(CM), and Unseen-Domain (UD)
		\item iii) \textbf{Numbers} denotes the datasets used for training and evaluation: ASVspoof19 (\circled{1}), FF++ (\circled{2}), ASVspoof21 (\circled{3}), WaveFake (\circled{4}), ITW (\circled{5}), DFDC (\circled{6}), CelebDF (\circled{7}), WildDF (\circled{8}), DF-1.0 (\circled{9}), DFFD (\circled{10}), FFIW (\circled{11}), DF40 (\circled{12}), DiffusionFace (\circled{13}), UADFV (\circled{14}), FakeAVCeleb (\circled{15}), KoDF (\circled{16}), DF-TIMIT (\circled{17}), MLAAD (\circled{18}), and ADD (\circled{19}).
	\end{tablenotes}
	\vspace*{-1.7\baselineskip}
\end{table}

\begin{figure}[ht]
	\centering
	\includegraphics[width=0.99\linewidth]{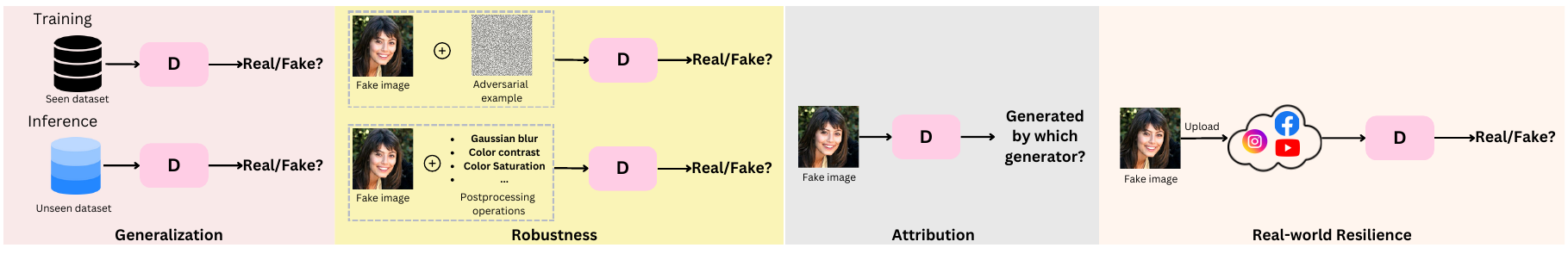}
	\caption{Illustration of passive DF detection approaches beyond detection, including: Generalization, Robustness, Attribution, and Interpretability.}
	\label{fig:beyond-detection}
	\vspace*{-1.7\baselineskip}
\end{figure}

\section{Real-world Requirements of Effective Passive DF Detection}\label{sec:beyond}
While achieving high detection accuracy on standard datasets represents a crucial first step, deploying effective DF detection systems in real-world scenarios requires addressing several additional challenges. Most current approaches demonstrate impressive performance within controlled settings but often fail when confronted with novel manipulation techniques, adversarial attacks, or DF quality degradation not represented in training data \cite{chen2022ost, hou2023evading, yan2024df40, xu2024profake}. This section explores critical aspects beyond accuracy that determine the practical utility of DF detection systems: generalization, robustness, attribution, and real-world adaptability. Figure \ref{fig:beyond-detection} illustrates these aspects.

\subsection{Generalization}
Generalization refers to the capability of DF detectors to maintain performance when confronted with previously unseen data distributions, manipulation techniques, and generator architectures. This capability is crucial as DF generation techniques continuously evolve. Researchers typically assess generalization through four protocols: (i) Within-domain evaluation, (ii) Cross-domain evaluation, (iii) Cross-manipulation evaluation, and (iv) Unknown-domain evaluation (Sec. \ref{ssec:evaluation}). We define 6 key approaches to enhance generalization, including data augmentation, synthetic data training, disentanglement learning, unsupervised learning, designed training strategies, and adaptive learning. Table \ref{tab:generalization} summarizes key ideas and evaluation protocols of existing methods.

\textbf{Data augmentation (DA) approaches} enhance model generalization by expanding training data diversity. Rather than collecting additional samples, these methods apply various transformations to existing data, creating a broader distribution of examples for the model to learn from. Depending on specific modalities, augmentation operations are various, including (i) image modality: Gaussian blur, JPEG compression, random crop \cite{ni2022core, luo2023beyond}; (ii) video modality: temporal dropout, temporal repeat, clip-level blending \cite{wang2023altfreezing}; (iii) audio modality: band-pass filters, time- and pitch-shifting, RawBoost \cite{doan2024balance, martin2024exploring}. Other works employ more sophisticated DA techniques, including dynamically erasing sensitive facial regions \cite{wang2021representative, das2021towards} to prevent overfitting and latent space augmentation to simulate variations within forgery features \cite{yan2023transcending, huang2025generalizable}. Contrastive learning (CL) has been integrated with DA to eliminate task-irrelevant contextual factors \cite{xu2022supervised, sun2022dual}.

\textbf{Synthetic data training methods} mitigate the dependence on available DF datasets by generating synthetic training examples during the learning process. \citeauthor{chen2022self} \cite{chen2022self} leverages the generator-discriminator framework with self-supervised auxiliary tasks to encourage learning generalizable features. In contrast, other works produce more challenging forgery artifacts by re-synthesizing real samples \cite{doan2024balance} or self-blending transformed versions of real samples \cite{shiohara2022detecting, yan2024generalizing}. \citeauthor{hasanaath2025fsbi} \cite{hasanaath2025fsbi} extends the idea of self-blending by extracting frequency domain features from self-blended images to amplify artifact information. Meanwhile, FreqBlender \cite{hanzhefreqblender} directly generates pseudo-fake faces in the frequency domain by using a decoder to estimate the probability map of the corresponding frequency knowledge.

\textbf{Disentanglement learning techniques} separate content-specific information (e.g., identity, background) from manipulation artifacts, allowing models to focus specifically on learning manipulation-related features without content-specific biases. \citeauthor{liang2022exploring} \cite{liang2022exploring} designs a dual-encoder architecture that separately extracts content and artifact features and employs CL to maximize the separation between these feature spaces. \citeauthor{yan2023ucf} \cite{yan2023ucf} proposes a multi-task disentanglement framework with a contrastive regularization loss for encouraging separation between common and specific forgery features while \citeauthor{guo2023controllable} \cite{guo2023controllable} constructs a controllable geometric embedding space to decouple irrelevant correlations. DiffusionFake \cite{chen2025diffusionfake} forces the detection network to learn disentangled representations of source and target features inherent in DFs by leveraging a pre-trained Stable Diffusion model to reconstruct both source and target identities. This technique helps the model to identify mixed identity information inherent in DFs.

\textbf{Unsupervised learning methods} leverage unlabeled data to learn more generalizable representations. Without relying on explicit labels which are time-consuming to collect, these approaches focus on learning universal patterns that distinguish authentic from manipulated content. To eliminate the need for pixel-level forgery annotations, UPCL \cite{zhuang2022uia} uses multivariate Gaussian estimation to model the distribution of real and fake image patches in a face region. Unsupervised contrastive learning (UCL) has been employed in \cite{qiao2024fully} to model the feature space based on intrinsic data characteristics, where similar samples are pulled together while dissimilar samples are pushed apart. Several researchers have explored transfer learning from self-supervised representations on generic datasets before fine-tuning for DF detection \cite{haliassos2022leveraging, kim2024one, oorloff2024avff}.

\textbf{Designed training strategy methods} structure the learning process to better capture fundamental differences between real and synthetic content. Rather than treating DF detection as a simple binary classification problem, they incorporate inductive biases about the underlying structure of manipulation artifacts into the training process, helping models learn more abstract, generalizable representations. \citeauthor{choi2024exploiting} \cite{choi2024exploiting} propose a two-stage training approach that first learns a robust style representation through supervised CL and then integrates these with content features for DF detection. This idea is extended by \citeauthor{zhu2025slim} \cite{zhu2025slim} for the audio modality. \citeauthor{cheng2024can} \cite{cheng2024can} introduce a progressive training strategy that organizes the latent feature space in a progressive manner, establishing a transition from real to fake as following a specific path: Real $\rightarrow$ Blendfake (self-blended) $\rightarrow$ Blendfake (cross-blended) $\rightarrow$ DF. This strategy helps the model to learn a more structured representation of forgery artifacts rather than treating each type as disconnected examples.

\textbf{Adaptive learning approaches} enable DF detectors to continuously adapt to new manipulation methods, generator families, or data distributions without compromising performance on previously learned knowledge. Common used techniques include meta-learning \cite{sun2021domain, han2023sigma, zhu2024tmfd, chen2022ost}, incremental learning \cite{pan2023dfil}, knowledge distillation \cite{lu2024one}, and domain generalization \cite{xie2023domain}. We refer readers to publications \cite{hospedales2021meta, van2022three, gou2021knowledge, zhou2022domain} for a better understanding of these techniques. Instead of fine-tuning the entire model, some studies design an adaption module to train a few parameters of detectors on new types of fake samples, such as leveraging low-rank adaptation matrices \cite{zhang2023adaptive, wu2024adapter} and designing a domain alignment loss on the source and target domain \cite{seraj2024semi}. Similarly, \citeauthor{oiso2024prompt} \cite{oiso2024prompt} uses prompt tuning on the target domain to adapt pre-trained models to new target domains by adding learnable prompts to the intermediate feature vectors in the front-end.


\begin{table}[ht]
	\centering
	\fontsize{5pt}{5pt}\selectfont
	\caption{A Summarization of Robustness Methods}
	\label{tab:robustness}
	\begin{tblr}{
			width = \linewidth,
			colspec = {Q[140]Q[35]Q[180]Q[35]Q[35]Q[35]Q[120]},
			cells = {c},
			cell{1}{1} = {r=2}{},
			cell{1}{2} = {r=2}{},
			cell{1}{3} = {r=2}{},
			cell{1}{4} = {c=4}{0.2\linewidth},
			cell{3}{1} = {r=6}{},
			cell{9}{1} = {r=6}{},
			vline{2-4} = {1-14}{},
			hline{1,15} = {-}{0.08em},
			hline{2} = {4-7}{},
			hline{3,9} = {-}{},
			hline{3} = {2}{-}{},
		}
		\textbf{Approach }                               & \textbf{Article }                  & \textbf{Key Idea }               & \textbf{Evaluation} &             &             &                       \\
		&                                    &                                  & \textbf{WA}         & \textbf{BA} & \textbf{TA} & \textbf{PA}           \\
		{Robustness to \\postprocessing attacks}            & \cite{devasthale2022adversarially} & Purely adversarial training      & \usym{2714}         & -           & -           & -                     \\
		& \cite{nguyen2024d}                 & Purely adversarial training      & -                   & -           & \usym{2714} & -                     \\
		& \cite{jeong2022frepgan}            & Frequency Perturbation           & \usym{2714}         & -           & -           & -                     \\
		& \cite{kawa2022defense}             & Adaptive adversarial training    & \usym{2714}         & -           & \usym{2714} & -                     \\
		& \cite{hooda2024d4}                 & Disjoint frequency ensemble      & -                   & \usym{2714} & -           & -                     \\
		& \cite{khan2024adversarially}       & Adversarial similarity loss      & \usym{2714}         & \usym{2714} & -           &                       \\
		{Robustness to \\postprocessing attacks     } & \cite{wang2023altfreezing}         & Data augmentation                & -                   & -           & -           & GB, BW, CC, C         \\
		& \cite{luo2023beyond}               & Data augmentation                & -                   & -           & -           & BW, C, GN, GB, CS, CC \\
		& \cite{yan2023transcending}         & Data augmentation                & -                   & -           & -           & GB, C, CC, CS, BW     \\
		& \cite{xu2024profake}               & Generator + progressive training & -                   & -           & -           & GB, C, R, GN          \\
		& \cite{feng2023self}                & Anomaly detection                & -                   & -           & -           & BW, C, GN, GB, CS, CC \\
		& \cite{chen2022ost}                 & Test-time training               & -                   & -           & -           & Unknown postprocessing
	\end{tblr}
	\begin{tablenotes}
		\item i) \textbf{Evaluation}: Protocols that methods evaluated on: White-box Attack (WA), Block-box Attack (BA), TA (Transferable Attack), Postprocessing Attack (PA)
		\item ii) \textbf{Postprocessing techniques}: Compression (C), Gaussian Blur (GB), Gaussian Noise (GN), Block Wise (BW), Change Contrast (CC), Change Saturation (CS), Resize (R)
	\end{tablenotes}
	\vspace*{-1.7\baselineskip}
\end{table}

\subsection{Robustness}
Robustness reflects the detector's ability to maintain reliable performance under various attack scenarios. We identify two critical threats that can compromise DF detectors' resilience: adversarial attacks and postprocessing attacks. Table \ref{tab:robustness} summarizes existing methods to improve robustness of DF detectors.

\textbf{Threat landscape.} DF detectors are increasingly vulnerable to adversarial attacks, which introduce carefully crafted perturbations into DF content to deceive detection systems while remaining imperceptible to humans. Recent studies have demonstrated that even SoTA detection models can be significantly compromised by adversarial examples (AEs) \cite{li2021exploring, jia2022exploring, neekhara2021adversarial, hou2023evading}. In contrast to adversarial attacks, postprocessing attacks employ various post-processing operations to alter DF quality, including compression, resizing, noise addition, and color adjustments. \cite{corvi2023intriguing, xu2024profake} show these techniques effectively obscure generation artifacts, substantially reducing detection performance. These attacks are particularly concerning as they often mirror common media processing operations, making them difficult to distinguish from benign transformations. Therefore, it is crucial to develop methods that enhance the detectors' robustness to adversarial attacks and postprocessing attacks.

\textbf{Robustness to adversarial attacks.} Adversarial training \cite{bai2021recent} is widely adopted to improve model robustness against adversarial attacks by incorporating adversarial examples into the training process. This technique has been demonstrated to be effective across both visual and audio modalities \cite{devasthale2022adversarially, nguyen2024d}. Beyond conventional adversarial training, more sophisticated approaches have recently emerged, including adaptive adversarial training that dynamically adjusts training samples based on attack difficulty \cite{kawa2022defense} and frequency-based adversarial training that uses a generator to generate frequency-level perturbation maps \cite{jeong2022frepgan}. For the black-box setting, D4 \cite{hooda2024d4} presents an ensemble-based method using multiple detector models to exam disjoint parts of the frequency spectrum, forcing attackers to find perturbations effective across multiple frequency ranges. \citeauthor{khan2024adversarially} \cite{khan2024adversarially} designs an adversarial similarity loss that maintains feature space proximity between original samples and their adversarial versions. These two techniques reduce the space of possible attacks and make successful attacks much harder to generate.

\textbf{Robustness to postprocessing attacks.} One straightforward approach is to simulate diverse postprocessing techniques during the training process. Multiple studies have demonstrated that data augmentation significantly improves detection robustness by exposing models to various postprocessing techniques they might encounter in real-world scenarios \cite{luo2023beyond, wang2023altfreezing, yan2023transcending}. Beyond simple augmentation, \citeauthor{xu2024profake} \cite{xu2024profake} combines a generator with progressive learning strategy to train the models from easy scenarios (non-degraded samples) to increasingly difficult ones (heavily-degraded samples). This method helps the model build robust representations while avoiding the training instability that can occur when immediately introducing heavily degraded examples. \citeauthor{feng2023self} \cite{feng2023self} cast the problem as anomaly detection where both original DFs and their degraded versions are considered anomalous deviations from authentic content. However, these methods work under the assumption that postprocessing techniques are well-known, which cannot adopt real-world settings where adversaries can apply unknown techniques. To mitigate this, OST \cite{chen2022ost} applies the test-time-training technique \cite{liu2021ttt++} that enables detectors to adapt to unknown techniques at inference time.

\begin{table}[ht]
	\centering
	\fontsize{5pt}{5pt}\selectfont
	\caption{A Summarization of Attribution Methods}
	\label{tab:attribution}
	\begin{tblr}{
			width = \linewidth,
			colspec = {Q[75]Q[35]Q[250]Q[250]Q[35]Q[35]Q[35]Q[35]},
			cells = {c},
			cell{1}{1} = {r=2}{},
			cell{1}{2} = {r=2}{},
			cell{1}{3} = {r=2}{},
			cell{1}{4} = {r=2}{},
			cell{1}{5} = {c=4}{0.15\linewidth},
			cell{3}{1} = {r=7}{},
			cell{10}{1} = {r=3}{},
			vline{2-5} = {1-12}{},
			hline{1,13} = {-}{0.08em},
			hline{2} = {5-8}{},
			hline{3,10} = {-}{},
			hline{3} = {2}{-}{},
		}
		\textbf{Approach }       & \textbf{Article }                            & \textbf{Key Idea }                                                                 & \textbf{Output }                                                                      & \textbf{Evaluation } &              &              &              \\
		&                                              &                                                                                    &                                                                                       & \textbf{MD}          & \textbf{MGM} & \textbf{CGM} & \textbf{UGM} \\
		Supervised Attribution   & \cite{marra2019gans}                         & Hand-crafted noise residuals                                                       & Real or Fake, GM type                                                        & \usym{2714}          & \usym{2714}  & \usym{2718}  & \usym{2718}  \\
		& \cite{yu2019attributing}                     & Learning-based fingerprint estimation                                              & Real or Fake, GM type                                                        & \usym{2714}          & \usym{2714}  & \usym{2718}  & \usym{2718}  \\
		& \cite{yang2022deepfake}                      & Apply image transformation to force the model focus on~architecture-related traces & Real or Fake, GM type                                                        & \usym{2714}          & \usym{2714}  & \usym{2714}  & \usym{2718}  \\
		& \cite{bui2022repmix}                         & Apply image transformation to force the model focus on~architecture-related traces & Real or Fake, GM type                                                        & \usym{2714}          & \usym{2714}  & \usym{2718}  & \usym{2718}  \\
		& \cite{klein2024source}                       & Learning-based audio fingerprint estimation                                        & Real or Fake, GM type                                                        & \usym{2714}          & \usym{2714}  & \usym{2718}  & \usym{2718}  \\
		& \cite{bhagtani2024attribution}               & Learning-based audio fingerprint estimation                                        & Real or Fake, GM type                                                        & \usym{2714}          & \usym{2714}  & \usym{2718}  & \usym{2714}  \\
		& \cite{xie2024generalized}                    & Two-stage audio training                                                           & Real or Fake, GM type                                                        & \usym{2714}          & \usym{2714}  & \usym{2718}  & \usym{2714}  \\
		Unsupervised Attribution & \cite{girish2021towards}                     & Clustering using K-means algorithm                                                 & Real or Fake, GM type                                                        & \usym{2714}          & \usym{2714}  & \usym{2718}  & \usym{2714}  \\
		& \cite{sun2023contrastive, sun2024rethinking} & Clustering through similarity matching~on visual features and pseudo-labeling      & Real or Fake, GM type                                                        & \usym{2714}          & \usym{2714}  & \usym{2718}  & \usym{2714}  \\
		& \cite{asnani2023reverse}                     & Model parsing + reverse engineering                                                & Real or Fake, GM type, network architecture, loss function, model parameters & \usym{2714}          & \usym{2714}  & \usym{2714}  & \usym{2714}
	\end{tblr}
	\begin{tablenotes}
		\item i) \textbf{Output} of the classifier
		\item ii) \textbf{Evaluation}: Settings to evaluate attribution methods: Multiple Data (MD), Multiple Known GMs (MGM), Changing GMs (CGM), and Unknown GMs (UGM)
	\end{tablenotes}
	\vspace*{-2.7\baselineskip}
\end{table}

\subsection{Attribution}
Attribution refers to identifying which specific GM was used to create a fake image, rather than just classifying an image as real or fake. This property provides deeper insights into the source and generation process of manipulated media. When malicious DFs are discovered, knowing which model created them helps investigators trace their origin and potentially identify patterns of coordinated misinformation campaigns. Attribution methods are evaluated under 4 key settings: (1) Multiple Data (MD), which tests performance across diverse datasets; (2) Multiple Known GMs (MGM), which assesses the ability to attribute images to their correct source among a set of known GMs; (3) Changing GMs (CGM), which evaluate on the known GMs but changing seeds, loss functions or datasets; and (4) Unknown GMs (UGM), which test on previously unseen GMs.

\textbf{Principle behind attribution methods.} DFs can be attributed to their source GMs through distinctive fingerprints embedded during the creation process \cite{wu2024traceevader}. These forensic traces manifest in two primary components: (i) High-frequency component traces, which contain architecture-specific signatures resulting from convolution filtering and upsampling operations; and (ii) Low-frequency component traces, which reflect instance-specific characteristics derived from unique model weights and initialization seeds.

Based on the knowledge requirement about the sources, attribution methods can be categorized into: supervised and unsupervised attribution. These methods are summarized in Table \ref{tab:attribution}.

\textbf{Supervised attribution methods} rely on prior knowledge of potential generator models and manually label each image with its source. These approaches train classifiers to distinguish between specific generator architectures by learning their characteristic fingerprints or artifacts. \citeauthor{marra2019gans} \cite{marra2019gans} extracted hand-crafted noise residuals from images to estimate unique fingerprints of each generator's architecture. \citeauthor{yu2019attributing} \cite{yu2019attributing} extended this idea to learning-based fingerprint estimation, which means that a classifier is trained on image-source pairs to learn fingerprints from images that are unique to each seen GAN model. To force the classifier to focus on architecture-related traces, other works propose to train the classifier on transformed images through applying different augmentation techniques \cite{yang2022deepfake, bui2022repmix}. The idea of supervised training has also been successfully applied in the audio modality for tracing fake audio back to their sources \cite{klein2024source, xie2024generalized, bhagtani2024attribution}.

\textbf{Unsupervised attribution methods} address the more challenging open-world scenario where new, previously unknown generation techniques may emerge. These approaches do not require explicit labels identifying the source of each sample, instead focusing on discovering intrinsic patterns or fingerprints that differentiate between different generation processes. \citeauthor{girish2021towards} \cite{girish2021towards} started with a small labeled set from known sources to train an initial network which is then used to extract features and cluster unlabeled images using K-means algorithm. This idea is extended by \cite{sun2023contrastive, sun2024rethinking} to improve clusters through similarity matching based on visual features and pseudo-labeling. \citeauthor{asnani2023reverse} \cite{asnani2023reverse} first constructes a network to estimate fingerprints left by the generators on their images and then a parsing network is designed to predict network architecture and loss function parameters from the estimated fingerprints.

\begin{table}[ht]
	\centering
	\fontsize{5pt}{5pt}\selectfont
	\caption{A Summarization of Real-world Resilience Methods}
	\label{tab:resilience}
	\begin{tblr}{
			width = \linewidth,
			colspec = {Q[70]Q[35]Q[230]Q[35]Q[170]Q[85]Q[35]Q[35]},
			cells = {c},
			cell{1}{1} = {r=3}{},
			cell{1}{2} = {r=3}{},
			cell{1}{3} = {r=3}{},
			cell{1}{4} = {c=5}{0.2\linewidth},
			cell{2}{4} = {c=2}{0.15\linewidth},
			cell{2}{6} = {r=2}{},
			cell{2}{7} = {c=2}{0.1\linewidth},
			cell{4}{1} = {r=6}{},
			cell{4}{3} = {r=3}{},
			cell{4}{4} = {r=3}{},
			cell{4}{5} = {r=3}{},
			cell{4}{6} = {r=3}{},
			cell{4}{7} = {r=3}{},
			cell{4}{8} = {r=3}{},
			cell{10}{1} = {r=2}{},
			vline{2-4, 6-7} = {1-12}{},
			hline{1,13} = {-}{0.08em},
			hline{2} = {4-8}{},
			hline{3} = {4-5,7-8}{},
			hline{4,10,12} = {-}{},
			hline{4} = {2}{-}{},
		}
		\textbf{\textbf{Challenge}} & \textbf{Article }         & \textbf{Key Idea }                                 & \textbf{Evaluation }  &                                                                       &                        &                                   &                                  \\
		&                           &                                                    & \textbf{Compression } &                                                                       & \textbf{Input Sizes }  & \textbf{Time Delay }              &                                  \\
		&                           &                                                    & \textbf{IC alg.}      & \textbf{AC alg.}                                                      &                        & \textbf{\textbf{Sequence length}} & \textbf{\textbf{Maximum offset}} \\
		Natural degradation         & \cite{woo2022add}         & Different models for different quality levels      & H.264 c23  c40        & -                                                                     & -                      & -                                 & -                                \\
		& \cite{lee2022bznet}       &                                                    &                       &                                                                       &                        &                                   &                                  \\
		& \cite{liao2023famm}       &                                                    &                       &                                                                       &                        &                                   &                                  \\
		& \cite{le2023quality}      & A unified model with collaborative learning        & H.264 c23  c40        & -                                                                     & 56, 112, 128, 224, 336 & -                                 & -                                \\
		& \cite{chen2024compressed} & 3D spatiotemporal trajectories~                    & H.264 c23  c40        & -                                                                     & -                      & -                                 & -                                \\
		& \cite{wang2024ftdkd}      & Knowledge distillation                             & ~-~                   & MP3, MP2, AAC, OGG, GSM, OPUS, AC3, DTS, WMA, RA~\cite{wang2024ftdkd} & -                      & -                                 & -                                \\
		Different input sizes       & \cite{xu2023tall}         & Using augmentation to simulate thumbnail           & -                     & -                                                                     & 112, 224               & -                                 & -                                \\
		& \cite{xu2024learning}     & Using augmentation to simulate thumbnail           & -                     & -                                                                     & 224, 448, 672          & -                                 & -                                \\
		Time delay                  & \cite{feng2023self}       & Time delay estimation through autoregressive model & -                     & -                                                                     & -                      & \usym{2714}                       & \usym{2714}
	\end{tblr}
	\begin{tablenotes}
		\item i) \textbf{Compression}: Algorithms used to simulate compression in real-world. Image Compression algorithms (IC alg.) and Audio Compression algorithms (AC alg.).
		\item ii) \textbf{Time delay}: The method is evaluated on different sequence length and maximum offset delay between frames and audio.
	\end{tablenotes}
	\vspace*{-1.7\baselineskip}
\end{table}

\subsection{Real-world Resilience.}
Real-world resilience represents a critical dimension of DF detection systems that focuses on maintaining reliable performance under naturally occurring real-world conditions. While robustness addresses deliberate adversarial and postprocessing attacks, real-world resilience concerns the detector's ability to function effectively despite routine transformations media undergoes without malicious intent. \circled{1} First, content uploaded to social media platforms (Instagram, TikTok, YouTube) or messaging applications (WhatsApp, Telegram) undergoes automatic compression with varying quality settings based on bandwidth considerations, device specifications, and platform requirements. For instance, YouTube employs different compression algorithms for different resolution options, while WhatsApp significantly reduces image quality during transmission. \circled{2} Second, detection systems must handle wide variations in media presentation. Videos uploaded to TikTok may be automatically fitted with thumbnails that resize content with different aspect ratios, potentially cropping key areas of the frame. Similarly, Instagram Reels limit video length and apply format-specific processing. In audio contexts, platforms like Instagram or Facebook limit clip duration, forcing content to be truncated. \circled{3} Third, videos in the real world often suffer from natural misalignment between their audio and visual streams due to technical issues in encoding and recording processes. A common example is when there is a consistent shift (time delay) of a few frames between what is seen and what is heard throughout the video. This creates a significant challenge for multimodal detectors since they cannot simply flag videos where the audio and visuals do not perfectly sync up. These transformations can inadvertently degrade detection performance by altering or removing subtle manipulation artifacts that detectors rely on.

One straightforward way to address the first problem is to employ multiple models for different quality levels \cite{liao2023famm, woo2022add, lee2022bznet}. However, these methods have significant limitations: (i) Cause computational costs and training data overhead and (ii) Not practical for real-world settings because they require prior knowledge about input video quality to select the appropriate detection model. \citeauthor{le2023quality} \cite{le2023quality} address these gaps by proposing a universal intra-model collaborative learning framework that enables effective simultaneous detection of different quality DFs using a single model. Differently, \citeauthor{chen2024compressed} \cite{chen2024compressed} uses 3D spatiotemporal trajectories to analyze facial landmark movements across frames with the hypothesis that video compression does not significantly alter the distribution of facial landmarks. In the audio modality, FTDKD \cite{wang2024ftdkd} uses the knowledge distillation technique to help the student model learn lost high-frequency information from the teacher model during compression. Regarding the second problem, the works by \cite{xu2023tall, xu2024learning} simulate the thumbnails by using augmentation techniques that mask fixed-size square areas at the same positions within frames. To handle the third problem in multimodal detectors, \citeauthor{feng2023self} \cite{feng2023self} utilizes an autoregressive model to estimate the temporal offset between each video frame and its corresponding audio signal, effectively capturing the time delay distribution. These methods are summarized in Table \ref{tab:resilience}.

\subsection{Discussion.}
\textbf{Generalization.} Table \ref{tab:generalization} reveals an imbalance in evaluation protocols across generalization methods. Most approaches focus on within-domain (WD) and cross-domain (CD) scenarios, with significantly fewer addressing cross-manipulation (CM) generalization and only a small fraction examining unknown-domain (UD) scenarios. This imbalance indicates that research has primarily focused on adapting to different datasets rather than novel manipulation techniques or generator architectures - a critical limitation given the rapid evolution of DF technology. Additionally, he predominant use of outdated training datasets like FF++ and ASVspoof19 likely constrains generalization capabilities. This pattern suggests that newer datasets representing emerging generation techniques (such as DF40 and VoiceWukong) should be utilized for training to better prepare detection systems for emerging generation techniques. \textbf{Robustness.} From Table \ref{tab:robustness}, we can observe that current approaches have largely focused on addressing either adversarial attacks or postprocessing attacks in isolation. This indicates that current approaches may fail to address real-world scenarios where both attack types may occur simultaneously. Moreover, only one work \cite{chen2022ost} directly addresses the challenge of unknown post-processing techniques, highlighting a critical research gap in adaptive robustness. \textbf{Attribution.} Table \ref{tab:attribution} reveals that supervised approaches perform well when identifying known GMs but struggle when confronted with changing generative architectures or entirely unknown models. Furthermore, only the work by \cite{asnani2023reverse} provides more information about the sources, including loss functions, model architectures, and model parameters, representing a substantial advancement in attribution depth and specificity. \textbf{Real-world Resilience.} From Table \ref{tab:resilience}, we can observe that most current approaches rely on simulation-based evaluations rather than testing on authentic platform-processed content. This gap between simulated conditions and actual platform-specific transformations may result in methods that perform well in controlled environments but fail when deployed in real applications where the specific transformations might differ significantly from those simulated during development.

However, trade-offs may exist if DF detection systems excel across all dimensions. \circled{1} \textbf{Trade-off between generalization and robustness.} Several works have theoretically and empirically demonstrated a trade-off between a DL model's ability to generalize well to unseen data (accuracy) and its ability to withstand small perturbations in the input (robustness) \cite{li2025triangular, zhang2019theoretically, stutz2019disentangling, gowda2024conserve}. This means that as we enhance a model's robustness against adversarial attacks, its standard generalization often decreases. Researchers should consider this trade-off and evaluate their robustness methods across datasets to assess the generalization. \circled{2} \textbf{Relationship between generalization and attribution.} Attribution requires detectors to retain and analyze fine-grained, generator-specific artifacts that might affect their generalization capability. \circled{3} \textbf{Relationship between robustness and attribution.} The study \cite{wu2024traceevader} reveals the vulnerability of existing attribution methods to postprocessing techniques that can disrupt specific traces and fingerprints left by GMs. This exploration raises a question about whether robustness approaches to postprocessing attacks can be employed to develop a robust attribution system.  \circled{4} \textbf{Computational efficiency considerations.} Approaches that attempt to satisfy these multiple aspects simultaneously might incur significant computational overhead, making them impractical for real-time applications or resource-constrained environments.

\section{Challenges and Future Directions} \label{sec:challenge-future}
\subsection{Detection of DFs in challenging conditions}
Current approaches often focus on fully synthetic content and demonstrate limited effectiveness in identifying DFs under challenging conditions, including face occlusion, multiple facial DFs, and partial manipulations.

\textbf{Face occlusion}. Advanced techniques, such as 3D face restoration \cite{chen2024blind} and blind inpainting \cite{criminisi2004region}, have successfully addressed occlusion challenges in generic face recognition. There exists significant potential to adapt and integrate these techniques into DF detection frameworks to maintain performance when faces are partially obscured or occluded.

\textbf{Multiple facial DFs}. Researchers evaluate their detection methods on specialized datasets designed for multi-subject forgery detection, including OpenForensics \cite{le2021openforensics} and DF-Platter \cite{narayan2023df}. These datasets specifically address the increasingly common scenario where multiple subjects within a single frame/image have been manipulated, presenting distinct challenges beyond single-face DF detection.

\textbf{Partial manipulation}. Detection systems must evolve to identify localized manipulations that alter only specific facial regions, partial frames, or partial utterances. Developing patch-based architectures VisionTransformer \cite{khan2022transformers} and implementing region-specific attention mechanisms could improve the detection of these increasingly sophisticated partial manipulations.

\subsection{Generalization capability}
\textbf{Generalization to unknown domains.} As shown in Table \ref{tab:generalization}, few methods are evaluated in the critical unknown-domain scenarios, in which test samples come from entirely unseen manipulation techniques, generators, and data distributions. Researchers should evaluate their methods in this real-world scenario with new datasets, such as DF40 \cite{yan2024df40}, VoiceWukong \cite{yan2024voicewukong}, and MLAAD \cite{muller2024mlaad}.

\textbf{Generalization at inference time.} To enhance the generalization of DF detectors to different data distributions, current approaches require detector retraining or access to labeled datasets \cite{chen2022ost, zhang2023adaptive, pan2023dfil, han2023sigma}, which presents significant practical limitations in real-world applications. In practice, test samples are always drawn from unknown data distributions; before new samples can be collected for training, detectors should classify these test samples. Recent advances in test-time adaptation \cite{liang2025comprehensive} and domain generalization \cite{wang2022generalizing} offer promising alternatives, enabling models to dynamically update parameters during inference without retraining or label access. These techniques should be considered to enhance the generalization of DF detectors at inference time.

\textbf{Generation and robustness trade-off.} Research from the broader computer vision field suggests that this trade-off cannot be entirely eliminated but can be strategically managed through architectural innovations and training strategies specifically tailored to the DF detection context \cite{gowda2024conserve, stutz2019disentangling}.

\subsection{Class imbalance issue}
DF detectors trained on current datasets face significant performance degradation when deployed in real-world settings due to class imbalance issues \cite{layton2024sok}. Particularly, these detectors exhibit a bias toward classifying content as fake when confronted with realistic deployment scenarios where authentic content vastly outnumbers DFs. This imbalance leads to excessive false positive rates that undermine the practical utility of detection systems. Long-tailed recognition is a potential direction to improve the accuracy of the "real" class with the least influence on the "fake" class \cite{miao2024out, bai2023effectiveness}. Additionally, researchers should consider reformulating DF detection as an anomaly detection problem where models learn a representation of authentic content variation rather than decision boundaries between real and fake classes \cite{ho2024long, salehi2021unified}. This approach naturally accommodates imbalanced scenarios by focusing on characterizing normal content patterns.

\subsection{Computation efficiency}
Many advanced detection approaches incur significant computational overhead, making them impractical for real-time or resource-constrained applications.

\textbf{Large-scale SSL front-ends in audio modality.} \citeauthor{wu2024adapter} \cite{wu2024adapter} indicates that using large pre-trained model as front-end cause computation overhead. Knowledge distillation \cite{gou2021knowledge} should be employed to transfer critical representations from these sophisticated models (teacher) to more compact DF classifiers (student), significantly reducing inference computational requirements while maintaining detection performance.

\textbf{Computational Considerations for multimodal detection.} Despite their effectiveness, multimodal approaches introduce significant computational overhead compared to unimodal methods. Processing both audio and visual streams simultaneously requires substantially more computational resources, presenting implementation challenges for real-time applications and resource-constrained environments. Parameter-efficient fine-tuning methodologies \cite{han2024parameter} offer promising solutions by adapting only a small subset of model parameters while keeping most pre-trained weights frozen, dramatically reducing both training and inference computational requirements. This techniques should be considered to employ for DF detection.

\subsection{Privacy-preserving detection}
Current detection approaches often require access to full media content or datasets for training, raising significant privacy concerns when deployed at scale across communication platforms or personal content. Federated learning \cite{zhang2021survey} is a distributed machine learning paradigm that enables multiple models to collaboratively train a global model without sharing their raw data. This framework should be adapted for DF detection to enable detector training and updating without centralizing sensitive media. Furthermore, researchers should consider developing privacy-preserving feature extraction techniques that convert media into non-reversible representations before analysis. These approaches would enable detection systems to operate on transformed data that preserves manipulation artifacts while removing personally identifiable information (e.g., speech content or identity information).

\section{Conclusion} \label{sec:conclusion}

This survey has presented a comprehensive analysis of passive DF detection approaches across image, video, audio, and multimodal modalities. We systematically categorized existing methods, examined their technical foundations, and highlighted their inherent strengths and limitations. Our work extends beyond conventional accuracy-focused evaluations to address critical requirements for real-world deployment: generalization capability, adversarial robustness, attribution precision, and real-world resilience. We identified significant limitations in current datasets and benchmarks, particularly their inadequate representation of real-world scenarios and prevalent class imbalance issues. Our analysis revealed key research gaps, including detection under challenging conditions (occlusions, multiple subjects), efficient generalization at inference time, and privacy-preserving detection frameworks.

As DF technology continues to evolve, detection methods must adapt accordingly. Future research should focus on developing models that maintain performance across novel generation techniques, demonstrate robustness against adversarial attacks and postprocssing techniques, and operate effectively under diverse real-world conditions. This multifaceted approach is essential for addressing the growing societal threats posed by increasingly sophisticated synthetic media.




\begin{acks}
	This publication has emanated from research conducted with the financial support of Science Foundation Ireland under Grant number 18/CRT/6183.
\end{acks}

\bibliographystyle{ACM-Reference-Format}
\bibliography{passive_reference_ACM}

%

\end{document}